\documentclass[12pt]{content/styles}

\usepackage{fullpage}
\usepackage[utf8]{inputenc} % allow utf-8 input
\usepackage[T1]{fontenc}    % use 8-bit T1 fonts
\usepackage{hyperref}       % hyperlinks
\usepackage{url}            % simple URL typesetting
\usepackage{booktabs}       % professional-quality tables
\usepackage{amsfonts}       % blackboard math symbols
\usepackage{nicefrac}       % compact symbols for 1/2, etc.
\usepackage{microtype}      % microtypography
\usepackage{xcolor}
\usepackage{graphicx}
\usepackage{subfigure} 
\usepackage{multirow}
\usepackage{amsmath}
\usepackage{amssymb, bm}
\usepackage{tablefootnote}
\usepackage{lineno}
\usepackage{pdflscape}
\usepackage{tabularx}
\usepackage{caption}

\usepackage{mathrsfs}
\usepackage{tipa}
\usepackage{verbatim}
\usepackage{makecell}

\usepackage{float}
\usepackage{color}
\usepackage{bbm}
\usepackage[linesnumbered,ruled]{algorithm2e}
\usepackage{algorithmic}
\usepackage{multicol}
\usepackage{wrapfig}
\usepackage{array}
\usepackage{setspace}
\usepackage{enumitem}
\usepackage{soul}
\usepackage{minitoc}
\usepackage{lineno}

\newcommand{\ourmodel}{CoD}

\begin{document} 
% \linenumbers

\begin{frontmatter}

\title{Continual Diffuser (CoD): Mastering Continual Offline Reinforcement Learning with Experience Rehearsal}

\author{Jifeng Hu${}^{1\dag}$\footnotetext{$^{\dag}$\text{Equal contribution.}}\footnotetext{$^{*}$\text{Correspondence: \url{chenhc@jlu.edu.cn}, \url{yichang@jlu.edu.cn}, and \url{dacheng.tao@gmail.com}.}}
\footnotetext{This work has been submitted to the IEEE for possible publication. Copyright may be transferred without notice, after which this version may no longer be accessible.}
}
\author{~~Li Shen${}^{2\dag}$}
\author{~~Sili Huang${}^{3\dag}$}
\author{~~Zhejian Yang${}^4$}
\author{~~Hechang Chen${}^5$$^{*}$}
\author{~~Lichao Sun${}^6$}
\author{~~Yi Chang${}^7$$^{*}$}
\author{~~Dacheng Tao${}^8$}

\address{${}^{\bm{1,3,4,5,7}}$School of Artificial Intelligence, Jilin University, Changchun, China \\
${}^{\bm{2}}$Sun Yat-sen University, Zhongshan, China \\
${}^{\bm{6}}$Lehigh University, Bethlehem, Pennsylvania, USA \\
${}^{\bm{8}}$College of Computing and Data Science, NTU, Singapore \\
}
\address{\url{https://github.com/JF-Hu/Continual_Diffuser}}

\begin{abstract}

Artificial neural networks, especially recent diffusion-based models, have shown remarkable superiority in gaming, control, and QA systems, where the training tasks' datasets are usually static.
However, in real-world applications, such as robotic control of reinforcement learning (RL), the tasks are changing, and new tasks arise in a sequential order.
This situation poses the new challenge of plasticity-stability trade-off for training an agent who can adapt to task changes and retain acquired knowledge.
In view of this, we propose a rehearsal-based continual diffusion model, called \underline{\textbf{Co}}ntinual \underline{\textbf{D}}iffuser (\textbf{\ourmodel{}}), to endow the diffuser with the capabilities of quick adaptation (plasticity) and lasting retention (stability).
Specifically, we first construct an offline benchmark that contains $90$ tasks from multiple domains.
Then, we train the \ourmodel{} on each task with sequential modeling and conditional generation for making decisions.
Next, we preserve a small portion of previous datasets as the rehearsal buffer and replay it to retain the acquired knowledge. 
Extensive experiments on a series of tasks show \ourmodel{} can achieve a promising plasticity-stability trade-off and outperform existing diffusion-based methods and other representative baselines on most tasks.
Source code is available at \href{https://github.com/JF-Hu/Continual_Diffuser}{here}.
\end{abstract}

% Artificial neural networks, especially recent diffusion-based models, have shown remarkable superiority in gaming, control, and QA systems, where the training tasks' datasets are usually static. However, in real-world applications, such as robotic control of reinforcement learning (RL), the tasks are changing, and new tasks arise in a sequential order. This situation poses the new challenge of plasticity-stability trade-off for training an agent who can adapt to task changes and retain acquired knowledge. In view of this, we propose a rehearsal-based continual diffusion model, called Continual Diffuser (CoD), to endow the diffuser with the capabilities of quick adaptation (plasticity) and lasting retention (stability). Specifically, we first construct an offline benchmark that contains 90 tasks from multiple domains. Then, we train the CoD on each task with sequential modeling and conditional generation for making decisions. Next, we preserve a small portion of previous datasets as the rehearsal buffer and replay it to retain the acquired knowledge. Extensive experiments on a series of tasks show CoD can achieve a promising plasticity-stability trade-off and outperform existing diffusion-based methods and other representative baselines on most tasks.

\end{frontmatter}

\section{Introduction}
Artificial neural networks, such as diffusion models, have made impressive successes in decision-making scenarios, e.g., game playing~\cite{mnih2015human}, robotics manipulation~\cite{kaufmann2023champion}, and autonomous driving~\cite{almalioglu2022deep}.
However, in most situations, a new challenge of difficult adaption to changing data arises when we adopt the general strategy of learning during the training phase and evaluating with fixed neural network weights~\cite{dohare2024loss}.
Changes are prevalent in real-world applications when performing learning in games, logistics, and control systems.
A crucial step towards achieving Artificial General Intelligence (AGI) is mastering the human-like ability to continuously learn and quickly adapt to new scenarios over the duration of their lifetime~\cite{berariu2021study}.
Unfortunately, it is usually ineffective for current methods to simply continue learning on new scenarios when new datasets arrive.
They will show a dilemma between storing historical knowledge (stability) in their brains and adapting to environmental changes (plasticity)~\cite{zeng2019continual}.

Recently, we have noticed that diffusion probabilistic models (DPMs) have emerged as an expressive structure for tackling complex decision-making tasks such as robotics manipulation by formulating deep reinforcement learning (RL) as a sequential modeling problem~\cite{he2023diffusion, wang2022diffusion, kang2023efficient}. 
Although recent DPMs have shown impressive performance in robotics manipulation, they, however, usually focus on a narrow setting, where the environment is well-defined and remains static all the time~\cite{ajay2022conditional, yang2023continual}, just like we introduce above.  
In contrast, in real-world applications, the environment changes dynamically in chronological order, forming a continuous stream of data encompassing various tasks.
In this situation, it is challenging for the agents to contain historical knowledge (stability) in their brains and adapt to environmental changes (plasticity) quickly based on already acquired knowledge~\cite{anand2023prediction, yue2024t}.
Thus, a natural question arises: 
\vspace{-5pt}
\begin{center}
\emph{Can we incorporate DPMs' merit of high expression and concurrently endow DPMs the ability towards better plasticity and stability in continual offline RL?}
\end{center}
\vspace{-5pt}

Facing the long-standing challenge of plasticity-stability dilemma in continual RL, current studies of continual learning can be roughly classified into three categories.
Structure-based methods~\cite{wang2022dirichlet, zhang2023split, smith2023continual, mendez2022reuse, mallya2018packnet} propose the use of a base model for pertaining and sub-modules for each task so as to store separated knowledge and reduce catastrophic forgetting. 
Regularization-based methods~\cite{zhang2023dynamics, zhang2022catastrophic, kirkpatrick2017overcoming, kessler2020unclear, nguyen2017variational} propose using auxiliary regularization loss such as $L_2$ penalty, KL divergence, and weight importance to contain policy optimization and avoid catastrophic forgetting during training.
Rehearsal-based methods~\cite{smith2023closer, peng2023ideal, huang2024solving, rolnick2019experience, chaudhry2018efficient} are considered simple yet effective in alleviating catastrophic forgetting as rehearsal mimics the memory consolidation mechanism of hippocampus replay inside biological systems.
There are many strategies to perform rehearsal. For instance, a typical method is gradient projection~\cite{chaudhry2018efficient}, which contains the gradients from new data loss as close as to previous tasks, furthest preventing performance decrease.

Although these methods are effective for continual learning, they present limited improvement in continual offline RL because of extra challenges such as distribution shift and value uncertain estimation.
Recently, diffusion-based methods, such as DD and Diffuser~\cite{ajay2022conditional, kang2023efficient, wang2022diffusion, janner2022planning}, propose to resolve the above two extra challenges from sequential modeling and have shown impressive results in many offline RL tasks.
However, they concentrate solely on training a diffuser that can only solve one task, thus showing limitations in real-world applications where training datasets or tasks usually arrive sequentially.
Though recent works, such as MTDIFF~\cite{he2023diffusion}, consider diffusers as planner or data generators for multi-task RL, the problem setting of their work is orthogonal to ours.

In this view, we take one step forward to investigate diffusers with arriving datasets and find that recent state-of-the-art diffusion-based models suffer from catastrophic forgetting when new tasks arrive sequentially (See Section \ref{Catastrophic Forgetting of Diffuser} for more details.).
To address this issue, we propose ``\underline{\textbf{Co}}ntinual \underline{\textbf{D}}iffuser'' (\textbf{\ourmodel{}}), which endows the diffuser with the capabilities of quickly adapting to new tasks (plasticity) meanwhile retaining the historical knowledge (stability) with experience rehearsal.
First of all, to take advantage of the potential of diffusion models, we construct an offline RL benchmark that consists of $90$ tasks from multiple domains, such as Continual World (CW) and Gym-MuJoCo.
These continual datasets will be released to all researchers soon at the present stage, and we will actively maintain and progressively incorporate more datasets into our benchmark.
Based on the benchmark, we train our method on each task with sequential modeling of trajectories and make decisions with conditional generation in evaluation.
Then, a small portion of each previous task dataset is reserved as the rehearsal buffer to replay periodically to our model.
Finally, extensive experiments on a series of tasks show that \ourmodel{} can achieve a promising plasticity-stability trade-off and outperform existing diffusion-based models and other representative continual RL methods on most tasks.
In summary, our contributions are threefold: 
\begin{itemize}
    \item We construct a continual offline RL benchmark that contains 90 tasks in the current stage, and we will actively incorporate more datasets for all researchers.
    \item We investigate the possibility of integrating experience rehearsal and diffuser, then propose the Continual Diffuser (\ourmodel{}) to balance plasticity and stability.
    \item Extensive experiments on a series of tasks show that \ourmodel{} can achieve a promising plasticity-stability trade-off and outperform existing baselines on most tasks.
\end{itemize}

\begin{figure*}[t!]
% \hspace{-1em}
%\vspace{-0.5em}
 \begin{center}
 \includegraphics[angle=0,width=0.99\textwidth]{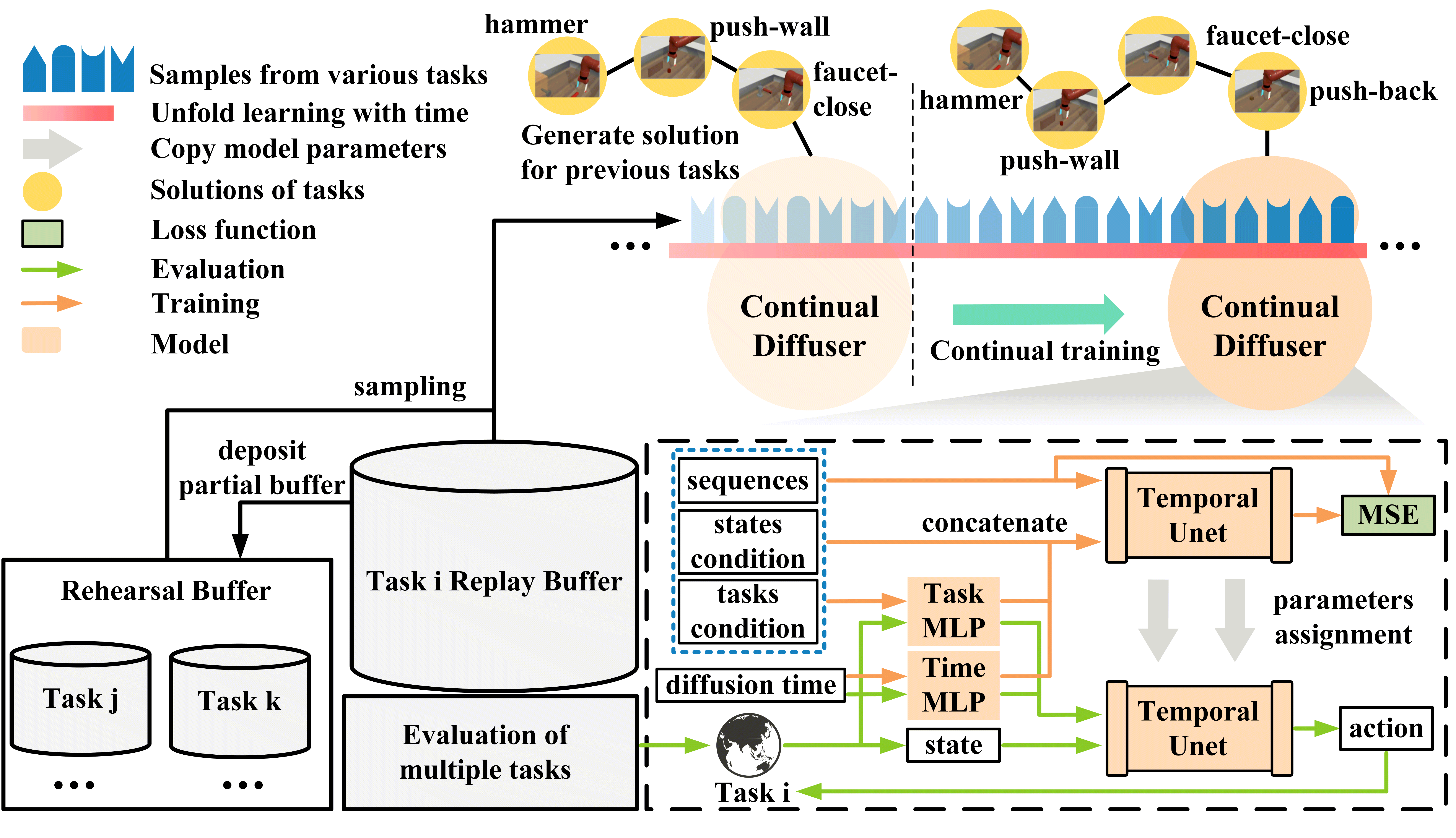}
 \caption{The framework of \ourmodel{}. Unfolding the training process with time, our model slides on the sample chain that is constructed by sampling from the current and rehearsal buffers. For each task $i$, \ourmodel{} replays small portion samples of previous tasks to reduce catastrophic forgetting and generate a solution that can solve all previous tasks. Detailed structure of \ourmodel{} is shown in the low right corner.}
 \label{framework}
 \end{center}
 \vspace{-0.3cm}
 % \vspace{-0.5em}
 \end{figure*}

\section{Results}

In this section, we will introduce environmental settings and evaluation metrics Section~\ref{Environmental Settings} and \ref{Evaluation Metrics}.
Then, in Section~\ref{Novel Benchmark for Continual Offline RL} and ~\ref{Baselines}, we first introduce a novel continual offline RL benchmark, including the task description and the corresponding dataset statistics, and introduce various baselines.
Finally, in Section~\ref{Results} and \ref{Ablation Study}, we report the comparison results, ablation study, and parameters sensitivity analysis.

\subsection{Environmental Settings}\label{Environmental Settings}
Following the same setting as prior works~\cite{zhang2023replay, yang2023continual}, we conduct thorough experiments on Continual World and Gym-MuJoCo benchmarks.
In Continual World, we adopt the task setting of CW10 and CW20 where CW20 means two concatenated CW10.
All CW tasks are version v1.
Besides, we also select Ant-dir for evaluation, which includes 40 tasks, and we arbitrarily select
four tasks (tasks-10-15-19-25) for training and evaluation. See Appendix~\ref{Statistics of Continual Offline RL Benchmarks} for more details.

\subsection{Evaluation Metrics}\label{Evaluation Metrics}
In order to compare the performance on a series of tasks, we follow previous studies~\cite{wolczyk2021continual, anand2023prediction} and adopt the totally average success rate $P(\rho)$ (higher is better), forward transfer $FT$ (higher is better), forgetting $F$ (lower is better), and the total performance $P+FT-F$ (higher is better) as evaluation metrics.
% These three metrics are canonical indicators used in the continual learning community.
Suppose that we use $p_i(\rho)$ to represent the average success rate on task $i$ at gradient update step $\rho$ and each task train $\Delta$ gradient steps, then the total average success rate $P(\rho)=\sum_{i=1}^{I}p_i(\rho)$, where $p_i(\rho) \in [0, 1]$.
The forward transfer $FT$ denotes the normalized AUC area between the training curve and the reference curve. 
Note that $FT_i<1$ and it might also be negative.
Mathematically, $FT=\frac{1}{I}\sum_{i}FT_i=\frac{1}{I}\sum_{i}\frac{AUC_i-AUC_{ref,i}}{1-AUC_{ref,i}}$, where we set $AUC_{ref,i}=0.5$ and $AUC_i=(p_i(i\cdot\Delta)+p_i((i+1)\cdot\Delta))/2$ for simplicity.
The forgetting $F_i$ is defined as the performance decrease between $p_i((i+1)\cdot\Delta)$ and $p_{I-1}(I\cdot\Delta)$, thus $F=\frac{1}{I}\sum_{i}^{I}F_i$.

\begin{figure*}[t!]
% \hspace{-1em}
 \begin{center}
 % \vspace{-0.5cm}
 \includegraphics[angle=0,width=0.65\textwidth]{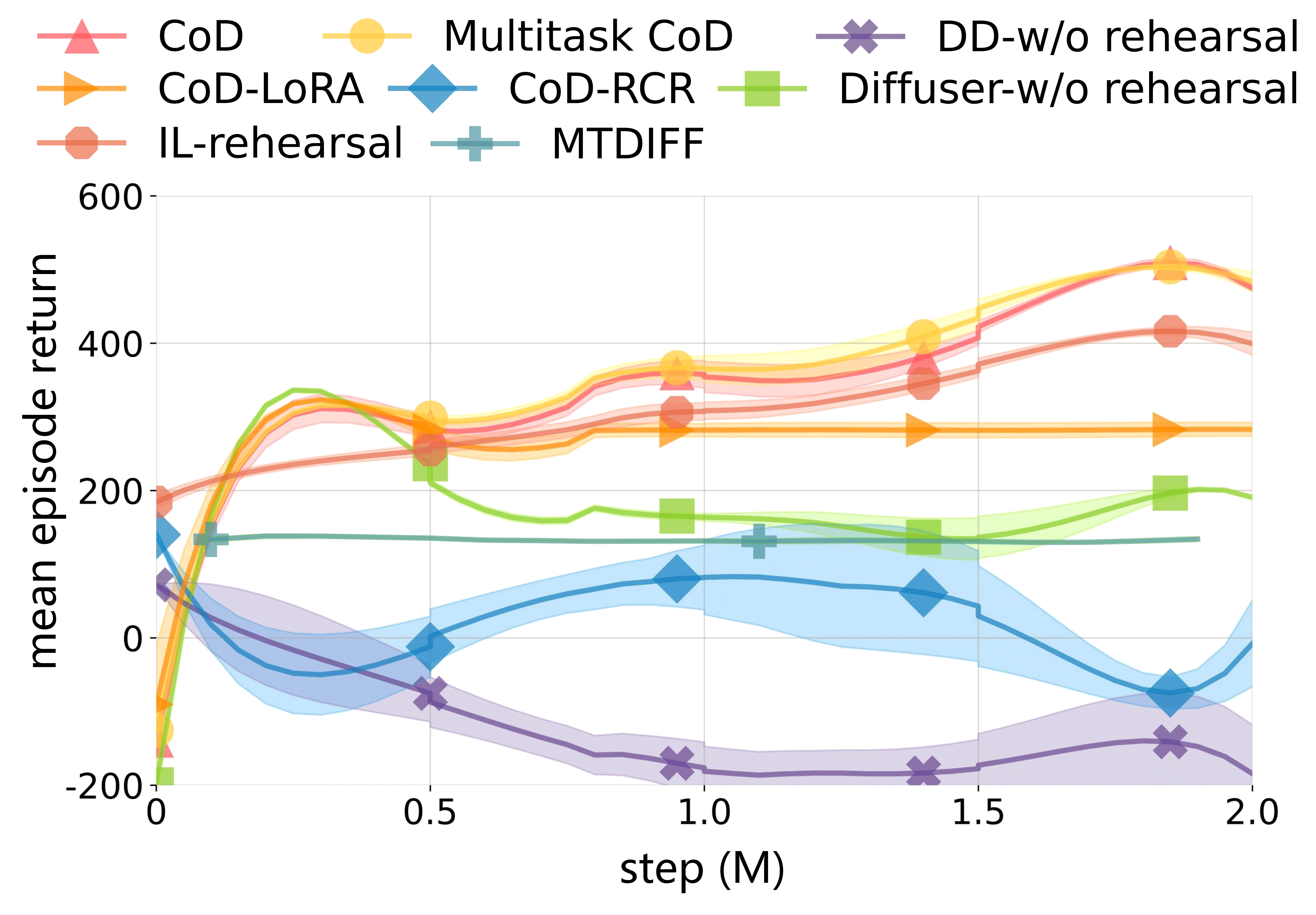}
 \caption{The comparison of \ourmodel{} and other diffusion-based models under the continual offline RL setting where ``w/o'' denotes ``without'', Multitask \ourmodel{} is a multitask variant of \ourmodel{}, \ourmodel{}-LoRA uses low-rank adaptation during training, and \ourmodel{}-RCR denotes that we train \ourmodel{} with return condition. IL-rehearsal denotes imitation learning with rehearsal. We train these methods on four arbitrarily selected tasks (tasks 10-15-19-25). The results show that previous diffusion-based methods (``DD-w/o rehearsal'', ``Diffuser-w/o rehearsal'', and ``MTDIFF'') exhibit severe forgetting when the datasets arrive sequentially.}
 \label{ant-dir online}
 \end{center}
 % \vspace{-0.3cm}
 \end{figure*}

\subsection{Novel Benchmark for Continual Offline RL}\label{Novel Benchmark for Continual Offline RL}

To take advantage of the potential of diffusion models, we propose a benchmark for continual offline RL (CORL), comprising datasets from 90 tasks, including 88 Continual World tasks and 2 Gym-MuJoCo tasks~\cite{wolczyk2021continual, todorov2012mujoco}. 
For the Gym-MuJoCo domain, there are 42 environmental variants, which are constructed by altering the agent goals.
In order to collect the offline datasets, we trained Soft Actor-Critic (SAC) on each task for approximately 1M time steps~\cite{haarnoja2018soft}.

Continual World~\cite{wolczyk2021continual} is a popular testbed that is constructed based on Meta-World~\cite{yu2020meta} and consists of realistic robotic manipulation such as Pushing, Reaching, and Door Opening.
CW is convenient for training and evaluating the abilities of forward transfer and forgetting because the state and action space are the same across all tasks.
Firstly, we will define the task-incremental CORL (TICORL), task-incremental CORL (TICORL), and task-incremental CORL (TICORL)~\cite{van2022three}.
In RL, we call the CL setting CICORL, where the CL tasks are constructed in the same environment with different goals, such as different directions or velocities.
We call the CL setting TICORL, where the CL tasks are indeed different environments but with the same purposes. For instance, the CL settings with the purpose of pushing blocks (e.g., ``push wall'' and ``push mug'' tasks in Continual World) in different robotic control tasks formulate the TICORL.
Finally, we can use the tasks of different purposes, such as push, pull, turn, and press blocks, to construct the DICORL.
For example, CW10 and CW20 form the mixed TICORL and DICORL setups because the task sequence contains multiple purposes. 
Additionally, Gym-MuJoCo's 42 environmental variants facilitate constructing a CICORL setup.
Researchers can use these datasets in any sequence or length for CL tasks to test the plasticity-stability trade-off of their methods. 
We also provide multiple quality datasets, such as `medium' and `expert,' in our benchmark.
We list the information statistics of our benchmark in Table~\ref{The information statistics of offline continual world datasets} and~\ref{The information statistics of offline continual world v2 datasets}, and Figure~\ref{continual world offline dataset return analysis} and~\ref{continual world-v2 offline dataset return analysis}, where the episodic time limit is set to 200, and the evaluation time step is set to 1M and 0.4M for different qualities datasets.

Ant-dir is an 8-joint ant environment. 
The different tasks are defined according to the target direction, where the agent should maximize its return with maximal speed in the pre-defined direction. 
As shown in Table~\ref{The information statistics of offline Gym-MuJoCo datasets}, there are 40 tasks (distinguished with ``task id'') with different uniformly sampled goal directions in Ant-dir. 
For each task, the dataset contains approximately 200k transitions, where the observation and action dimensions are 27 and 8, respectively.
We found that the Ant-dir datasets have been used by many researchers~\cite{xu2022prompting, li2020multi, rakelly2019efficient}, so we incorporate them into our benchmark.
Moreover, we report the mean return information of each sub-task in Table~\ref{The information statistics of offline Gym-MuJoCo datasets} and Figure~\ref{ant-dir offline dataset return analysis}.
As for Cheetah-dir, it only contains two tasks that represent forward and backward goal directions.
Compared with Ant-dir, Cheetah-dir possesses lower observation and action space.

\subsection{Baselines}\label{Baselines}
We compare our method (\ourmodel{}) with various representative baselines, encompassing structure-based, regularization-based, and rehearsal-based methods.
In structure-based methods, we select LoRA~\cite{li2023loftq}, PackNet~\cite{mallya2018packnet}, and Multitask.
For regularization-based methods, we select L2, EWC~\cite{kirkpatrick2017overcoming}, MAS~\cite{aljundi2018memory}, and VCL~\cite{nguyen2017variational} for evaluation.
Rehearsal-based baselines include t-DGR~\cite{yue2024t}, DGR~\cite{shin2017continual}, CRIL~\cite{gao2021cril}, A-GEM~\cite{chaudhry2018efficient}, and IL~\cite{ho2016generative}.
Besides, we also include several diffusion-based methods~\cite{janner2022planning, ajay2022conditional} and Multitask methods, such as MTDIFF~\cite{he2023diffusion} for the evaluation.

\begin{table*}[t!]
\centering
\small
\caption{The performance comparison on the Continual World and Ant-dir datasets. We compare our method (\ourmodel{}) with baselines trained with the offline pattern as well as the online pattern. We report the average success rate, backward forgetting, and forward transfer of our method and several representative baselines in Continual World tasks (shown in parts (a) and (b)). Moreover, we conduct experiments on CW4 ({``hammer-v1'', ``push-wall-v1'', ``faucet-close-v1'', ``push-back-v1''}) with mixed-quality datasets and show the results in part (c). For Ant-dir datasets shown in part (d), we report the comparison results with diffusion-based, non-diffusion-based, and multitask methods.}
% \vspace{0.5em}
\label{The performance comparison on offline and online}
\resizebox{\textwidth}{!}{
\begin{tabular}{l | l | r  r  r  r  r  r  r  r}
\toprule
\specialrule{0em}{1.5pt}{1.5pt}
\toprule
\multicolumn{2}{c|}{} & \multicolumn{4}{c|}{Continual World 10}  & \multicolumn{4}{c}{Continual World 20} \\
\midrule[1pt]
\makecell[c]{train\\mode} & Model & P $\uparrow$ & FT $\uparrow$ & F $\downarrow$ & \multicolumn{1}{c|}{P+FT-F $\uparrow$} & P $\uparrow$ & FT $\uparrow$ & F $\downarrow$ & P+FT-F $\uparrow$\\
\midrule[1pt]
\multirow{7}{*}{\makecell[c]{(a) offline\\baselines}} & EWC       & 0.20\tiny{$\pm$0.16} & 0.30\tiny{$\pm$0.21} & 0.80\tiny{$\pm$0.16} & \multicolumn{1}{r|}{-0.30} &  0.30\tiny{$\pm$0.21}& 0.30\tiny{$\pm$0.21} & 0.70\tiny{$\pm$0.21} & -0.10\\
 & Finetune  & 0.20\tiny{$\pm$0.16} & 0.10\tiny{$\pm$0.09} & 0.80\tiny{$\pm$0.16} & \multicolumn{1}{r|}{-0.50} & 0.10\tiny{$\pm$0.09}& 0.10\tiny{$\pm$0.09} & 0.80\tiny{$\pm$0.16} & -0.60\\
 & DGR       & 0.30\tiny{$\pm$0.21} & 0.90\tiny{$\pm$0.09} & 0.70\tiny{$\pm$0.21} & \multicolumn{1}{r|}{0.50} & 0.50\tiny{$\pm$0.25}& 0.90\tiny{$\pm$0.09} & 0.50\tiny{$\pm$0.25} & 0.90\\
 & t-DGR      & 0.40\tiny{$\pm$0.24} & 0.70\tiny{$\pm$0.21} & 0.60\tiny{$\pm$0.24} & \multicolumn{1}{r|}{0.50} & 0.50\tiny{$\pm$0.25}& 0.90\tiny{$\pm$0.09} & 0.50\tiny{$\pm$0.25} & 0.90\\
 & CRIL      & 0.70\tiny{$\pm$0.21} & 0.80\tiny{$\pm$0.16} & 0.20\tiny{$\pm$0.16} & \multicolumn{1}{r|}{1.30} & 0.70\tiny{$\pm$0.21}& 0.90\tiny{$\pm$0.09} & 0.00\tiny{$\pm$0.00} & 1.60\\
 & Multitask & 1.00\tiny{$\pm$0.00} & 0.90\tiny{$\pm$0.09} & 0.00\tiny{$\pm$0.00} & \multicolumn{1}{r|}{1.90} & 1.00\tiny{$\pm$0.00}& 0.90\tiny{$\pm$0.09} & 0.00\tiny{$\pm$0.00} & 1.90\\
 & \ourmodel{} & 0.98\tiny{$\pm$0.01} & 0.89\tiny{$\pm$0.09} & -0.01\tiny{$\pm$0.001} & \multicolumn{1}{r|}{1.88} & 0.98\tiny{$\pm$0.01} & 0.89\tiny{$\pm$0.09} & 0.00\tiny{$\pm$0.00} & 1.87\\
 \midrule[1pt]
 \midrule[1pt]
\multirow{8}{*}{\makecell[c]{(b) online\\baselines}} & A-GEM & 0.02\tiny{$\pm$0.01} & -0.76\tiny{$\pm$0.02} & 0.22\tiny{$\pm$0.02} & \multicolumn{1}{r|}{-0.96} & 0.17\tiny{$\pm$0.10}& 0.17\tiny{$\pm$0.11} & 0.64\tiny{$\pm$0.12} & -0.30\\
 & PackNet & 0.05\tiny{$\pm$0.01} & -0.60\tiny{$\pm$0.01} & 0.35\tiny{$\pm$0.02} & \multicolumn{1}{r|}{-0.90} & 0.14\tiny{$\pm$0.09}& -0.34\tiny{$\pm$0.19} & 0.53\tiny{$\pm$0.20} & -0.73\\
 & VCL & 0.10\tiny{$\pm$0.03} & -0.81\tiny{$\pm$0.05} & -0.02\tiny{$\pm$0.008} & \multicolumn{1}{r|}{-0.69} & 0.18\tiny{$\pm$0.13}& -0.62\tiny{$\pm$0.14} & 0.02\tiny{$\pm$0.06} & -0.46\\
 & MAS & 0.23\tiny{$\pm$0.05} & -0.63\tiny{$\pm$0.08} & -0.05\tiny{$\pm$0.03} & \multicolumn{1}{r|}{-0.35} & 0.41\tiny{$\pm$0.15}& -0.12\tiny{$\pm$0.18} & -0.01\tiny{$\pm$0.01} & 0.30\\
 & EWC & 0.30\tiny{$\pm$0.03} & -0.36\tiny{$\pm$0.05} & 0.02\tiny{$\pm$0.02} & \multicolumn{1}{r|}{-0.08} & 0.56\tiny{$\pm$0.20}& 0.13\tiny{$\pm$0.28} & 0.01\tiny{$\pm$0.02} & 0.68\\
 & L2 & 0.21\tiny{$\pm$0.03} & -0.58\tiny{$\pm$0.07} & 0.02\tiny{$\pm$0.02} & \multicolumn{1}{r|}{-0.39} & 0.51\tiny{$\pm$0.09}& 0.12\tiny{$\pm$0.19} & 0.10\tiny{$\pm$0.03} & 0.53\\
 & Multitask & 1.00\tiny{$\pm$0.00} & 0.90\tiny{$\pm$0.09} & 0.00\tiny{$\pm$0.00} & \multicolumn{1}{r|}{1.90} & 1.00\tiny{$\pm$0.00}& 0.90\tiny{$\pm$0.09} & 0.00\tiny{$\pm$0.00} & 1.90\\
 \midrule[1pt]
 \midrule[1pt]
\multicolumn{2}{c|}{} & \multicolumn{4}{c|}{Continual World 4}  & \multicolumn{4}{c}{}\\
\midrule[1pt]
 \multirow{2}{*}{\makecell[c]{(c) offline\\baselines}} & IL-rehearsal & 0.57\tiny{$\pm$0.19} & 0.12\tiny{$\pm$0.54} & 0.18\tiny{$\pm$0.09} & \multicolumn{1}{r|}{0.51} \\
 & CoD & 0.85\tiny{$\pm$0.02} & 0.60\tiny{$\pm$0.13} & 0.05\tiny{$\pm$0.01} & \multicolumn{1}{r|}{1.40} \\
 \midrule[1pt]
 \midrule[1pt]
\multicolumn{2}{c|}{} & \multicolumn{8}{c}{Ant-dir} \\
 \midrule[1pt]
 \multirow{3}{*}{\makecell[c]{(d) offline\\baselines}} & Model & \ourmodel{} & \makecell[r]{Multitask \\ \ourmodel{}} & \makecell[r]{IL-\\rehearsal} & \makecell[r]{\ourmodel{}-\\LoRA} & \makecell[r]{Diffuser-w/o \\ rehearsal} & \ourmodel{}-RCR & MTDIFF & \makecell[r]{DD-w/o \\ rehearsal} \\
  & \makecell[l]{Mean\\return} & 478.19\tiny{$\pm$15.84} & 485.15\tiny{$\pm$~~5.86} & 402.53\tiny{$\pm$17.67} & 296.03\tiny{$\pm$11.95} & 270.44\tiny{$\pm$~~5.54} & 140.44\tiny{$\pm$32.11} & 84.01\tiny{$\pm$41.10} & -11.15\tiny{$\pm$45.27} \\
\bottomrule
\specialrule{0em}{1.5pt}{1.5pt}
\bottomrule
\end{tabular}}
% \vspace{-0.3cm}
\end{table*}

\begin{figure*}[t!]
% \hspace{-1em}
 \begin{center}
 % \vspace{-0.5cm}
 \includegraphics[angle=0,width=0.99\textwidth]{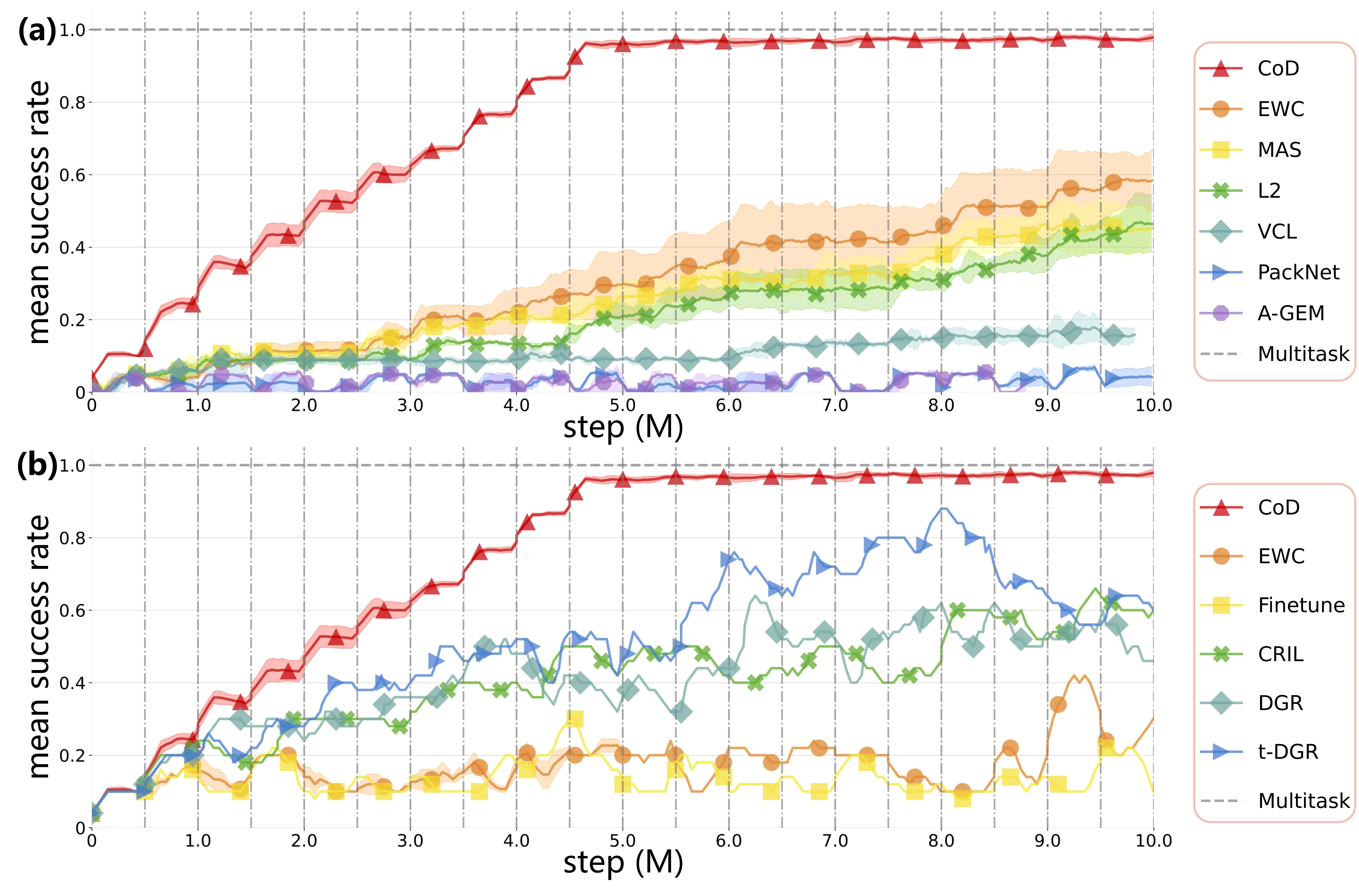}
 \caption{The comparison of our method \ourmodel{} and other baselines on CW20 where these baselines are trained with online and offline datasets and are trained with 500k gradient steps on each task. In the above figure, we use the dash-dotted lines to indicate the task changes. Part (a) shows the comparison where the baselines are trained in online mode, while in part (b), the baselines are trained with offline datasets.}
 \label{cw20 total success rate comparison of offline and online}
 \end{center}
 % \vspace{-0.3cm}
 \end{figure*}

\subsection{Main Results}\label{Results}
\noindent\textbf{Ant-dir Results.}~~~
To show the effectiveness of our method in reducing catastrophic forgetting, we compare our method with other diffusion-based methods on the Ant-dir tasks ordered by 10-15-19-25.
As shown in Table~\ref{The performance comparison on offline and online} (d) and Figure~\ref{ant-dir online}, the results illustrate: 
1) Directly applying previous diffusion-based methods into continual offline RL will lead to severe catastrophic forgetting because the scores of Diffuser-w/o rehearsal and DD-w/o rehearsal are far behind \ourmodel{}.
2) Extending the technique of LoRA into the diffusion model may not always work. The reason lies in that the parameter quantity size is small, which inspires us to construct diffuser foundation models in future work.
3) Rehearsal can bring significant improvements on diffuser as \ourmodel{} approaches the score of Multitask \ourmodel{}.

\noindent\textbf{Online Continual World Results.}~~~
Considering that offline datasets prohibit further exploration in the environments, which may hinder the capability of some baselines that are designed for online training.
We conduct CW10 and CW20 experiments of these methods under the online continual RL setting.
Similarly, we constrain the interaction as 500k time steps for each task and report the comparison results in Figure~\ref{cw20 total success rate comparison of offline and online} (a) and Table~\ref{The performance comparison on offline and online} (a).
The results show that our method (\ourmodel{}) surpasses other baselines by a large margin, which illustrates the superior performance over balancing plasticity and stability.
Besides, it is indeed that some methods, such as EWC, are more suitable for online training by comparing the performances in Figure~\ref{cw20 total success rate comparison of offline and online} (a) and (b). 
Additionally, we also report the comparison under mixed-quality datasets CL setting in Table~\ref{The performance comparison on offline and online} (c).
Please refer to Appendix~\ref{Additional Experiments} for the comparison of model plasticity and generation acceleration details.

\noindent\textbf{Offline Continual World Results.}~~~
This section presents the comparison between \ourmodel{} and six representative continual RL methods on CW10 and CW20 benchmarks.
In order to show the capabilities of plasticity (quick adaptation to unseen tasks) and stability (lasting retention of previous knowledge), we keep the size of training samples, number of gradient updates, and computation constant.
Figure~\ref{cw20 total success rate comparison of offline and online} (b) and Table~\ref{The performance comparison on offline and online} (b) summarize the results of CW10 and CW20 tasks.
We observe that our method can quickly master these manipulation tasks and remember the acquired knowledge when new tasks arrive, while the baselines (except for Multitask) struggle between plasticity and stability because the performance of these baselines fluctuates among tasks.
Moreover, after 5M gradient steps, our method still remembers how to solve the same task it learns, which shows small forgetting. 
The results of the table also show that though some baselines exhibit high forward transfer, the average success rate is lower than our method, and they forget knowledge fleetly.

 \begin{figure*}[t!]
% \hspace{-1em}
%\vspace{-0.5em}
 \begin{center}
 \includegraphics[angle=0,width=0.8\textwidth]{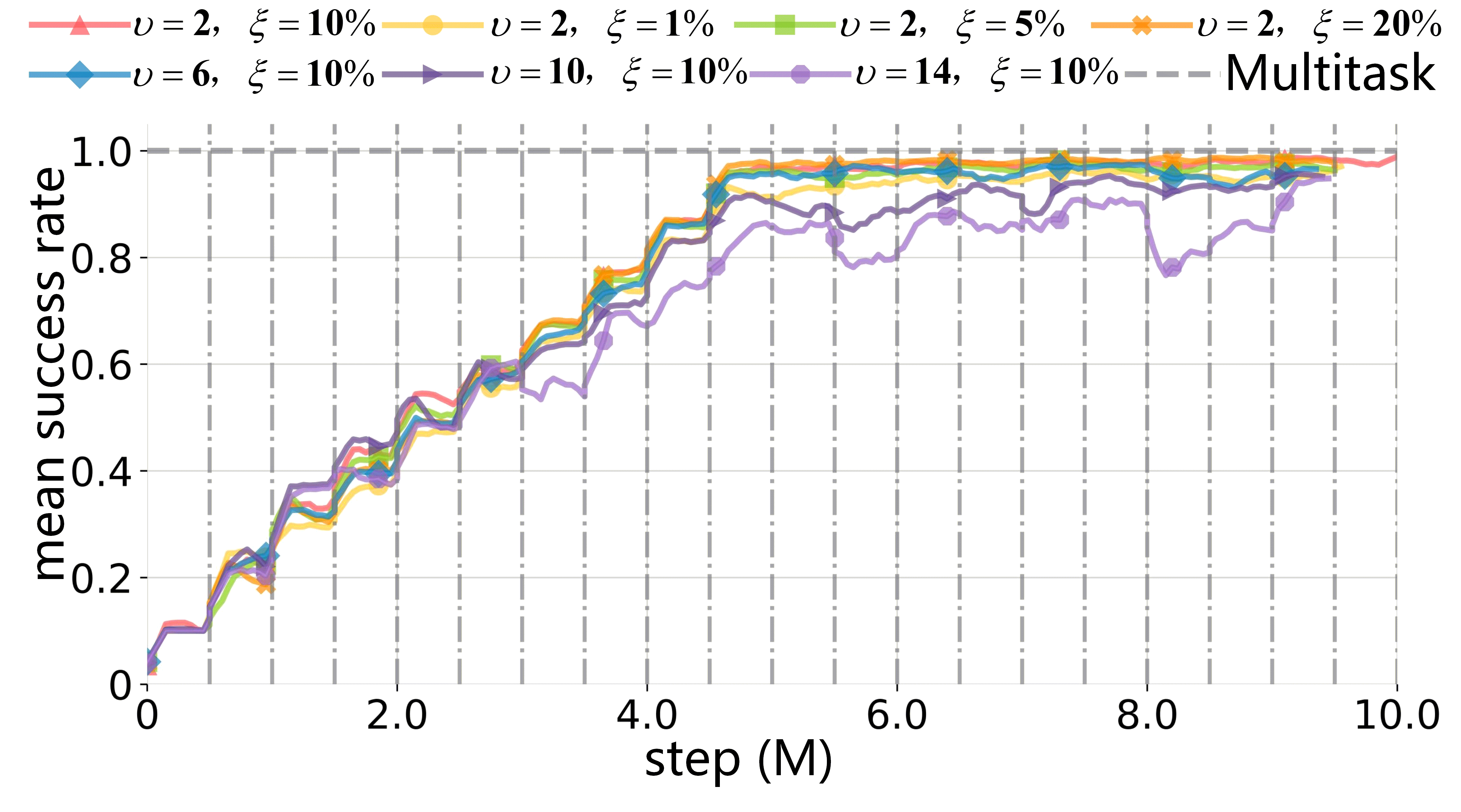}
 \caption{The parameters sensitivity analysis of rehearsal frequency $\upsilon$ and rehearsal sample diversity $\xi$ on CW20.}
 \label{cw20 ablation study}
 \end{center}
 \vspace{-0.5cm}
 \end{figure*}

\subsection{Ablation Study}\label{Ablation Study} 
To show the effectiveness of experience rehearsal, we conduct an ablation study of \ourmodel{} in CW and Ant-dir tasks.
We compare our method with and without experience rehearsal and find that experience rehearsal indeed brings significant performance gain. 
For example, \ourmodel{} achieves 76.82\% performance gain compared with \ourmodel{}-w/o rehearsal.
In CW 20 tasks, \ourmodel{} reaches mean success rate from 20\% to 98\% when incorporating experience rehearsal.
Refer to Table~\ref{ablation study} for more results.

\noindent\textbf{Sensitivity of Key Hyperparameters.}~~~
In the experiments, we introduce the key hyper-parameters: the rehearsal frequency ($\upsilon$) and rehearsal sample diversity ($\xi$).
The larger $\upsilon$ will aggravate the catastrophic forgetting because the model can access previous samples after a longer training process.
A large value of $\xi$ will improve the performance and increase the storage burden, while a small value is more cost-efficient for longer CL tasks but is more challenging to hold the performance.
We conduct the sensitivity of the hyperparameters on the CW and Ant-dir environments, and the results are shown in Figure~\ref{cw20 ablation study} and Figure~\ref{parameters sensitivity of Ant-dir}.
According to the results, our method can still reach good performance with the variation of $\upsilon$ and $\xi$.

\section{Discussion}

\subsection{Catastrophic Forgetting of Diffuser}\label{Catastrophic Forgetting of Diffuser}

Previous diffusion-based methods~\cite{ajay2022conditional, janner2022planning, yue2024t}, such as DD and Diffuser, are usually proposed to solve a single task, which is not in line with the real-world situation where the task will dynamically change.
Thus, it is meaningful but challenging to train a diffuser that can adapt to new tasks (plasticity) while retaining historical knowledge.
When we directly extend the original diffusion-based method in continual offline RL, we can imagine that severe catastrophic forgetting will arise in the performance because there are no mechanisms to retain preceding knowledge. 
As shown in Figure~\ref{ant-dir online}, in order to show the catastrophic forgetting, we compare our method and the representative diffusion-based methods on Ant-dir, where we arbitrarily select four tasks, task-10, task-15, task-19, and task-25, to form the CL setting.
Diffuser-w/o rehearsal and DD-w/o rehearsal represent the original method Diffuser and DD, respectively.
Multitask \ourmodel{} and MTDIFF are the multitask baselines, which can access all training datasets in any time step, and \ourmodel{}-RCR represents we use return condition for decision generation during the training stage.
\ourmodel{}-LoRA denotes that we train \ourmodel{} with the technique of low-rank adaptation.
IL-rehearsal is the imitation learning with rehearsal.
The results show that previous diffusion-based methods exhibit severe catastrophic forgetting when the datasets arrive sequentially, and at the same time, the good performance of \ourmodel{} illustrates experience rehearsal is effective in reducing catastrophic forgetting.

\subsection{Reducing Catastrophic Forgetting with Experience Rehearsal}\label{Reducing Catastrophic Forgetting with Experience Rehearsal}

In Section~\ref{Results}, we illustrate the effectiveness of experience rehearsal through the experiments on our proposed offline CL benchmark, which contains 90 tasks for evaluation.
From the perspective of the CL tasks quantity, we evaluate carious quantity settings, such as 4 tasks for Ant-dir, 4 tasks for CW4, 10 tasks for CW10, and 20 tasks for CW20.
From the perspective of classification of traditional CL settings, our experimental settings contain CICORL, TICORL, and DICORL.
In the Ant-dir environment, we select 10-15-19-25 task sequence as the CL setting and conduct the experiment compared with other diffusion-based methods.
From the results shown in Figure~\ref{ant-dir online}, we can see distinct catastrophic forgetting on the recent diffusion-based method, though they show strong performance in other offline RL tasks~\cite{kang2023efficient, he2023diffusion}.
To borrow the merits of diffusion models' strong expression on offline RL and equip them with the ability to reduce catastrophic forgetting, we propose to use experience rehearsal to master the CORL.
Detailed architecture is shown in Figure~\ref{framework}, and we postpone the method description in Section~\ref{Continual Diffuser}.

\begin{figure*}[t!]
% \hspace{-1em}
%\vspace{-0.5em}
 \begin{center}
 \includegraphics[angle=0,width=0.99\textwidth]{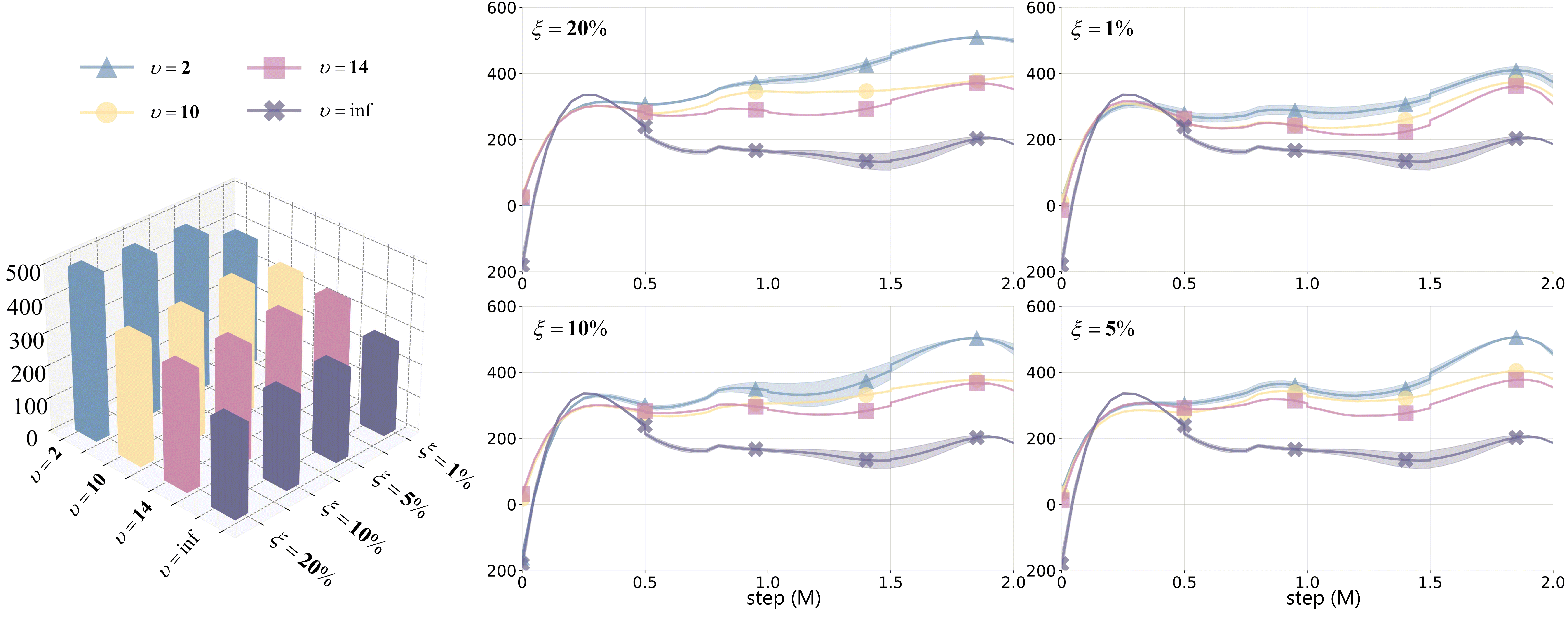}
 \caption{The parameters sensitivity of Ant-dir.}
 \label{parameters sensitivity of Ant-dir}
 \end{center}
 % \vspace{-0.3cm}
 % \vspace{-0.5em}
 \end{figure*}

Apart from the Ant-dir environment, we also report the performance on more complex CL tasks, i.e., CW10 and CW20, in Table~\ref{The performance comparison on offline and online}.
Considering that most baselines are trained in online mode in their original papers, we first select the online baselines and compare their mean success rate with our method.
The results (Table~\ref{The performance comparison on offline and online} and Figure~\ref{cw20 total success rate comparison of offline and online}) show that our method (\ourmodel{}) surpasses other baselines by a large margin, which illustrates the superior performance over balancing plasticity and stability.
Besides, we also compare our method with these baselines trained with offline datasets, where the results show that our method can quickly master these manipulation tasks and remember the acquired knowledge when new tasks arrive, while the baselines (except for Multitask) struggle between plasticity and stability because the performance of these baselines fluctuates among tasks.
When the previous tasks appear once again after 5M training steps, the baselines show different levels of catastrophic forgetting because the performance decreases after 5M steps. 
However, our method still remembers how to solve the same task it learned before, which shows small forgetting.
Moreover, we also conduct mixed-quality dataset experiments to show our method's capability of learning from sub-optimal offline datasets.
For more details, please refer to Appendix~\ref{Additional Experiments}.

To investigate the influence of key hyperparameters, we report the performance of the rehearsal frequency ($\upsilon$) and rehearsal sample diversity ($\xi$) in Figure~\ref{cw20 ablation study} and Figure~\ref{parameters sensitivity of Ant-dir}, where larger $\upsilon$ corresponds to aggravated catastrophic forgetting and a larger value of $\xi$ will improve the performance and increase the storage burden.
In practice, we find that usually $\upsilon=2$ and $\xi=10\%$ indicate good performance and pose small challenges for the computation and memory burden (see Appendix~\ref{Implementation Details} for memory and efficiency analysis.).

\section{Methods}

\subsection{Continual Offline RL}

In this paper, we focus on the task-incremental setting of task-aware continual learning in the offline RL field where the different tasks come successively for training~\cite{zhang2023replay, wang2023distributionally, smith2023continual, schwarz2018progress, abel2023definition, wang2023comprehensive}.
Each task is defined as a corresponding Markov Decision Process (MDP) $\mathcal{M}=\langle\mathcal{S}, \mathcal{A}, \mathcal{P}, \mathcal{R}, \gamma\rangle$, where $\mathcal{S}$ and $\mathcal{A}$ represent the state and action space, respectively, $\mathcal{P}: \mathcal{S}\times\mathcal{A}\rightarrow \Delta(\mathcal{S})$ denotes the Markovian transition probability, $\mathcal{R}: \mathcal{S}\times\mathcal{A}\times\mathcal{S}\rightarrow \mathbb{R}$ is the reward function, and $\gamma\in [0, 1)$ is the discount factor.
In order to distinguish different tasks, we use subscript $i$ for task $i$, such as $\mathcal{M}_{i}$, $\mathcal{S}_{i}, \mathcal{A}_{i}, \mathcal{P}_{i}, \mathcal{R}_{i}$, and $\gamma_{i}$.
At each time step $t$ in task $i$, the agent receives a state $s_{i,t}$ from the environment and produces an action $a_{i,t}$ with a stochastic or deterministic policy $\pi$. 
Then a reward $r_{i,t}=r(s_{i,t}, a_{i,t})$ from the environment serves as the feedback to the executed action of the agent.
Continual offline RL aims to find an optimal policy that can maximize the discounted return $\sum_i^{I}\mathbb{E}_{\pi}[\sum_{t=0}^{\infty}\gamma^{t}r(s_{i,t}, a_{i,t})]$~\cite{yang2023continual, sun2023smart, wei2023imitation} on all tasks with previously collected dataset $\{D_i\}_{i\in I}$.

\subsection{Conditional Diffusion Probabilistic Models}

In this paper, diffusion-based models are proposed to model the distribution of trajectory $\tau$, where each trajectory can be regarded as a data point.
Then we can use diffusion models to learn the trajectory distribution $q(\tau)=\int q(\tau^{0:K}) d\tau^{1:K}$ with a predefined forward diffusion process $q(\tau^{k}|\tau^{k-1})=\mathcal{N}(\tau^{k};\sqrt{\alpha_k}\tau^{k-1},(1-\alpha_k)\bm{I})$ and the trainable reverse process $p_{\theta}(\tau^{k-1}|\tau^{k})=\mathcal{N}(\tau^{k-1};\mu_{\theta}(\tau^{k},k), \Sigma_k)$, where $k\in [1, K]$ is the diffusion step, $\sqrt{\alpha_{k}}$ and $\sqrt{1-\alpha_{k}}$ control the drift and diffusion coefficients, $\mu_{\theta}(\tau^{k})=\frac{1}{\sqrt{\alpha_k}}(\tau_k-\frac{\beta_k}{\sqrt{1-\bar{\alpha}_k}}\epsilon_{\theta}(\tau^{k},k))$, $\Sigma_k=\frac{1-\bar{\alpha}_{k-1}}{1-\bar{\alpha}_{k}}\beta_{k}\bm{I}$, and $\alpha_k + \beta_k = 1$.
$\epsilon_{\theta}(\tau^k, k)$ represents the noising model~\citep{sohl2015deep}.
According to \cite{ho2020denoising}, we can train $\epsilon_{\theta}(\tau^k, k)$ with the below simplified objective 
\begin{equation*}
    \mathcal{L}(\theta)=\mathbb{E}_{k\sim U(1, 2, ..., K), \epsilon\sim\mathcal{N}(0,\bm{I}), \tau^0\sim D}[||\epsilon-\epsilon_{\theta}(\tau^k, k)||_2^2],
\end{equation*}
where $k$ is the diffusion time step, $U$ is uniform distribution, $\epsilon$ is multivariant Gaussian noise, $\tau^0=\tau$ is sampled from the replay buffer $D$, and $\theta$ is the parameters of model $\epsilon_{\theta}$. 

Conditions play a vital role in conditional generation because this method makes the outputs of diffusion models controllable.
We can also use two conditions methods, classifier-guided and classifier-free, to train diffusion models $p_{\theta}(\tau^{k-1}|\tau^{k}, \mathcal{C})$~\citep{liu2023more}.
The classifier-guided method separates the training of the unconditional diffusion model and conditional guide and then combines them together, i.e., $p_{\theta, \phi}(\tau^{k-1}|\tau^{k}, \mathcal{C})\propto p_{\theta}(\tau^{k-1}|\tau^{k})p_{\phi}(\mathcal{C}|\tau^{k})$.
The corresponding sampling process is $p(\tau^{k-1}|\tau^{k}, \mathcal{C})=\mathcal{N}(\mu_{\theta}+\Sigma_k \cdot \nabla log~p_{\phi}(\mathcal{C}|\tau), \Sigma_k)$.
Compared with classifier-guided, the classifier-free method implicitly builds the correlation between the trajectories and conditions in the training phase by learning unconditional and conditional noise $\epsilon_{\theta}(\tau^k, \emptyset, k)$ and $\epsilon_{\theta}(\tau^k, \mathcal{C}, k)$, where $\emptyset$ is usually the zero vector~\citep{ajay2022conditional}. 
Then the perturbed noise at each diffusion time step is calculated by $\epsilon_{\theta}(\tau^k, \emptyset, k)+\omega(\epsilon_{\theta}(\tau^k, \mathcal{C}, k)-\epsilon_{\theta}(\tau^k, \emptyset, k))$.
In this paper, we adopt the classifier-free guidance due to its simplicity, controllability, and higher performance~\cite{ajay2022conditional}.

\subsection{Continual Diffuser}\label{Continual Diffuser}

In this section, we introduce the Continual Diffuser (\ourmodel{}), as shown in Figure~\ref{framework}, which contains classifier-free task-conditional training, experience rehearsal, and conditional generation for decision.

\noindent\textbf{Data Organization.}~~~~
In RL, we leverage the characteristic of the diffusion model that can capture joint distributions in high-dimensional continual space by formulating the training data from single-step transition to multi-step sequences.
Specifically, we have $I$ tasks, and each task $\mathcal{M}_i$ consists of $N$ trajectories $\{\tau_i\}_{1}^{N}$, where the $\tau_{i,n}=\{s_{i,t,n}, a_{i,t,n}\}$ will be split into equaling sequences with $T_e$ time steps as the discrepancy of trajectories may occur across tasks.
In the following parts, we slightly abuse this notation $\tau_{i}$ to represent the sequence data with length $T_e$ sampled from task $i$ 's dataset $D_i$ and $\hat{\tau}_{i}$ to denote the generative sequence.

\noindent\textbf{Task Condition.}~~~
In order to distinguish different tasks, we propose to use environment-related information as the task condition.
For example, in the Ant-dir environment, the agent's goal is to maximize its speed in the pre-defined direction, which is given as the goal in the specific tasks.
So, we propose to use this information as condition $\mathcal{C}_{task}$ to train our model.
In each diffusion step $k$, the task condition $\mathcal{C}_{task}$ will pass through a task embedding function to obtain task embedding, which will be fed into the diffusion model jointly with diffusion time step embedding.  
Apart from the task conditions that are used implicitly in the training, we also need explicit observation conditions.
We use the first state $s_{i,t,n}$ of the $T_e$ length sampled sequence $\tau_{i,n} = \{s_{i,t,n}, a_{i,t,n}, s_{i,t+1,n}, a_{i,t+1,n}, ..., s_{i,t+T_e-1,n}, a_{i,t+T_e-1,n}\}$ as the conditions.
Then at each diffusion generation step, after we obtain the generated sequences $\{\hat{s}_{i,t,n}, \hat{a}_{i,t,n}, ..., \hat{s}_{i,t+T_e-1,n}, \hat{a}_{i,t+T_e-1,n}\}^k$, the first observation $\hat{s}_{i,t,n}$ is directly replaced by $s_{i,t,n}$, i.e., $\hat{\tau}_{i,n}^k = \{s_{i,t,n}, \hat{a}_{i,t,n}, ..., \hat{s}_{i,t+T_e-1,n}, \hat{a}_{i,t+T_e-1,n}\}^k$.

\noindent\textbf{Training Objective.}~~~~
Following the previous studies of the diffusion model~\cite{ho2020denoising, he2023diffusion}, the training and generation for each task $i$ are defined as 
\begin{equation}\label{single task DM loss}
    \mathcal{L}_{i}(\theta)=\mathbb{E}_{k\sim U(1, K), \epsilon\sim\mathcal{N}(0,\bm{I}), \tau^0_{i}\sim D_{i}}[||\epsilon-\epsilon_{\theta}(\tau_{i}^k,\mathcal{C}_{task~i}, k)||_2^2],
\end{equation}
\begin{equation}\label{generation equation}
    \tau_{i}^{k-1}=\frac{\sqrt{\bar{\alpha}_{k-1}}\beta_{k}}{1-\bar{\alpha}_k}\cdot\bar{\tau}_{i}+\frac{\sqrt{\alpha}_k (1-\bar{\alpha}_{k-1})}{1-\bar{\alpha}_{k}}\tau_{i}^{k}+|\Sigma_{k}|\bm{z},
\end{equation}
where $z\sim\mathcal{N}(\bm{0},\bm{I})$, $\bar{\tau}_{i}=\frac{\tau_{i}^{k}-\sqrt{1-\bar{\alpha}_k}\bar{\epsilon}}{\sqrt{\bar{\alpha}_k}}, |\Sigma_k|=\frac{1-\bar{\alpha}_{k-1}}{1-\bar{\alpha}_{k}}\beta_{k}$, and $\bar{\epsilon}=\epsilon_{\theta}(\tau_{i}^k, \emptyset, k)+\omega(\epsilon_{\theta}(\tau_{i}^k, \mathcal{C}_{task}, k)-\epsilon_{\theta}(\tau_{i}^k, \emptyset, k))$.

\noindent\textbf{Experience Rehearsal.}~~~
In this paper, we propose periodic rehearsal to strengthen the knowledge of previous tasks, which mimics the memory consolidation mechanism of hippocampus replay inside biological systems.
When a new dataset $D_i$ of task $i$ arrives, we preserve a small portion $\xi$ of the entire dataset, donated as $\mathscr{D}_i$.
For the small training dataset $\mathscr{D}_i$, it is easy to overfit these data for most rehearsal-based methods.
Fortunately, inspired by the distributional robust optimization, increasing the hardness of the samples will hinder memory overfitting.
The discrete type of diffusion process $\tau^{k}=\sqrt{\alpha_k}\tau^{k-1}+\sqrt{1-\alpha_k}\epsilon$ can be reformulated as the corresponding continuous forward process $d\tau=-\frac{1}{2}\beta(t)\tau dt+\sqrt{\beta(t)}dW$, where $W$ is the standard Wiener process (a.k.a. Brownian motion).
This process gradually inserts directional noise (i.e., increasing the hardness) to induce transformation from trajectory distribution to Gaussian distribution.
So rehearsal-based diffusers naturally possess the capability of reducing memory overfitting, and the total objective function is 
\begin{equation}\label{total loss func}
\small
    \min_{\forall\theta\in\Theta}[\mathbb{E}_{\tau_j\in D_j}\mathcal{L}_j(\theta,\tau_j,\mathcal{C}_{task~j})+\mathbb{E}_{\tau_i\in \mathscr{D}_i, i<j}\mathcal{L}_i(\theta,\tau_i,\mathcal{C}_{task~i})]
\end{equation}
In practice, we usually set the rehearsal frequency $\upsilon$ as 2 gradient steps and the portion $\xi$ as 10\%.

\noindent\textbf{Architecture.}~~~
In this paper, we adopt temporal Unet with one-dimensional convolution blocks as the diffusion model to predict noises.
Specifically, temporal Unet contains several down-sampling blocks, a middle block, several up-sampling blocks, a time embedding block, and a task embedding block.
We train the time embedding block and task embedding block to generate time and task embeddings that are added to the observation-action sequence
\begin{equation*}
    \tau_{i,t:t+T_e-1,n}=\begin{pmatrix}
        s_{i,t,n} & s_{i,t+1,n} & ... & s_{i,t+T_e-1,n}\\
        a_{i,t,n} & a_{i,t+1,n} & ... & a_{i,t+T_e-1,n}\\
    \end{pmatrix}.
\end{equation*}
In the return conditional diffusion models, we replace the task embedding block with the return embedding block.
Also, following the implementation of low-rank adaptation in Natural Language Processing~\cite{hu2021lora, peft}, we increase the LoRA module in down-sampling, middle, and up-sampling blocks to construct the LoRA variant \ourmodel{}-LoRA.

\subsection{Conclusion}\label{Conclusion and Limitations}

First of all, to facilitate the development of the continual offline RL community, a continual offline benchmark that contains 90 tasks is constructed based on Continual World and Gym-MuJoCo. 
Based on our benchmark, we propose Continual Diffuser (\ourmodel{}), an effective continual offline RL method that possesses the capabilities of plasticity and stability with experience rehearsal.
Finally, extensive experiments illustrate the superior plasticity-stability trade-off when compared with representative continual RL baselines.

\section*{AUTHOR CONTRIBUTION}
Jifeng Hu and Hechang Chen contributed to the dataset collection, figures, and tables. 
Li Shen, Sili Huang, and Jifeng Hu contributed to the article's organization.
The all authors contributed to the article's writing.

\section*{CODE AND DATA AVAILABILITY}
The code and data are available in GitHub at \url{https://github.com/JF-Hu/Continual_Diffuser}.

\section*{ACKNOWLEDGEMENT}
We would like to thank Lijun Bian for her contributions to the figures and tables of this manuscript.
We thank Runliang Niu for his contributions to providing help on the computing resource.

\bibliography{example_paper}

\begin{thebibliography}{83}
\providecommand{\natexlab}[1]{#1}
\providecommand{\url}[1]{\texttt{#1}}
\expandafter\ifx\csname urlstyle\endcsname\relax
  \providecommand{\doi}[1]{doi: #1}\else
  \providecommand{\doi}{doi: \begingroup \urlstyle{rm}\Url}\fi

\bibitem[Abel et~al.(2023)Abel, Barreto, Van~Roy, Precup, van Hasselt, and Singh]{abel2023definition}
David Abel, Andr{\'e} Barreto, Benjamin Van~Roy, Doina Precup, Hado van Hasselt, and Satinder Singh.
\newblock A definition of continual reinforcement learning.
\newblock \emph{arXiv preprint arXiv:2307.11046}, 2023.

\bibitem[Ajay et~al.(2022)Ajay, Du, Gupta, Tenenbaum, Jaakkola, and Agrawal]{ajay2022conditional}
Anurag Ajay, Yilun Du, Abhi Gupta, Joshua Tenenbaum, Tommi Jaakkola, and Pulkit Agrawal.
\newblock Is conditional generative modeling all you need for decision-making?
\newblock \emph{arXiv preprint arXiv:2211.15657}, 2022.

\bibitem[Aljundi et~al.(2018)Aljundi, Babiloni, Elhoseiny, Rohrbach, and Tuytelaars]{aljundi2018memory}
Rahaf Aljundi, Francesca Babiloni, Mohamed Elhoseiny, Marcus Rohrbach, and Tinne Tuytelaars.
\newblock Memory aware synapses: Learning what (not) to forget.
\newblock In \emph{Proceedings of the European conference on computer vision (ECCV)}, pages 139--154, 2018.

\bibitem[Almalioglu et~al.(2022)Almalioglu, Turan, Trigoni, and Markham]{almalioglu2022deep}
Yasin Almalioglu, Mehmet Turan, Niki Trigoni, and Andrew Markham.
\newblock Deep learning-based robust positioning for all-weather autonomous driving.
\newblock \emph{Nature machine intelligence}, 4\penalty0 (9):\penalty0 749--760, 2022.

\bibitem[Anand and Precup(2023)]{anand2023prediction}
Nishanth Anand and Doina Precup.
\newblock Prediction and control in continual reinforcement learning.
\newblock \emph{arXiv preprint arXiv:2312.11669}, 2023.

\bibitem[Atkinson et~al.(2021)Atkinson, McCane, Szymanski, and Robins]{atkinson2021pseudo}
Craig Atkinson, Brendan McCane, Lech Szymanski, and Anthony Robins.
\newblock Pseudo-rehearsal: Achieving deep reinforcement learning without catastrophic forgetting.
\newblock \emph{Neurocomputing}, 428:\penalty0 291--307, 2021.

\bibitem[Beeson and Montana(2023)]{beeson2023balancing}
Alex Beeson and Giovanni Montana.
\newblock Balancing policy constraint and ensemble size in uncertainty-based offline reinforcement learning.
\newblock \emph{arXiv preprint arXiv:2303.14716}, 2023.

\bibitem[Berariu et~al.(2021)Berariu, Czarnecki, De, Bornschein, Smith, Pascanu, and Clopath]{berariu2021study}
Tudor Berariu, Wojciech Czarnecki, Soham De, Jorg Bornschein, Samuel Smith, Razvan Pascanu, and Claudia Clopath.
\newblock A study on the plasticity of neural networks.
\newblock \emph{arXiv preprint arXiv:2106.00042}, 2021.

\bibitem[Chaudhry et~al.(2018)Chaudhry, Ranzato, Rohrbach, and Elhoseiny]{chaudhry2018efficient}
Arslan Chaudhry, Marc'Aurelio Ranzato, Marcus Rohrbach, and Mohamed Elhoseiny.
\newblock Efficient lifelong learning with a-gem.
\newblock \emph{arXiv preprint arXiv:1812.00420}, 2018.

\bibitem[Chen et~al.(2021)Chen, Lu, Rajeswaran, Lee, Grover, Laskin, Abbeel, Srinivas, and Mordatch]{chen2021decision}
Lili Chen, Kevin Lu, Aravind Rajeswaran, Kimin Lee, Aditya Grover, Misha Laskin, Pieter Abbeel, Aravind Srinivas, and Igor Mordatch.
\newblock Decision transformer: Reinforcement learning via sequence modeling.
\newblock \emph{Advances in neural information processing systems}, 34:\penalty0 15084--15097, 2021.

\bibitem[Chi et~al.(2023)Chi, Feng, Du, Xu, Cousineau, Burchfiel, and Song]{chi2023diffusion}
Cheng Chi, Siyuan Feng, Yilun Du, Zhenjia Xu, Eric Cousineau, Benjamin Burchfiel, and Shuran Song.
\newblock Diffusion policy: Visuomotor policy learning via action diffusion.
\newblock \emph{arXiv preprint arXiv:2303.04137}, 2023.

\bibitem[Dohare et~al.(2024)Dohare, Hernandez-Garcia, Lan, Rahman, Mahmood, and Sutton]{dohare2024loss}
Shibhansh Dohare, J~Fernando Hernandez-Garcia, Qingfeng Lan, Parash Rahman, A~Rupam Mahmood, and Richard~S Sutton.
\newblock Loss of plasticity in deep continual learning.
\newblock \emph{Nature}, 632\penalty0 (8026):\penalty0 768--774, 2024.

\bibitem[Fontanesi et~al.(2019)Fontanesi, Gluth, Spektor, and Rieskamp]{fontanesi2019reinforcement}
Laura Fontanesi, Sebastian Gluth, Mikhail~S Spektor, and J{\"o}rg Rieskamp.
\newblock A reinforcement learning diffusion decision model for value-based decisions.
\newblock \emph{Psychonomic bulletin \& review}, 26\penalty0 (4):\penalty0 1099--1121, 2019.

\bibitem[Foret et~al.(2020)Foret, Kleiner, Mobahi, and Neyshabur]{foret2020sharpness}
Pierre Foret, Ariel Kleiner, Hossein Mobahi, and Behnam Neyshabur.
\newblock Sharpness-aware minimization for efficiently improving generalization.
\newblock \emph{arXiv preprint arXiv:2010.01412}, 2020.

\bibitem[Gao et~al.(2021)Gao, Gao, Guo, Zhang, and Chen]{gao2021cril}
Chongkai Gao, Haichuan Gao, Shangqi Guo, Tianren Zhang, and Feng Chen.
\newblock Cril: Continual robot imitation learning via generative and prediction model.
\newblock In \emph{2021 IEEE/RSJ International Conference on Intelligent Robots and Systems (IROS)}, pages 6747--5754. IEEE, 2021.

\bibitem[Ghosh et~al.(2022)Ghosh, Ajay, Agrawal, and Levine]{ghosh2022offline}
Dibya Ghosh, Anurag Ajay, Pulkit Agrawal, and Sergey Levine.
\newblock Offline rl policies should be trained to be adaptive.
\newblock In \emph{International Conference on Machine Learning}, pages 7513--7530. PMLR, 2022.

\bibitem[Haarnoja et~al.(2018)Haarnoja, Zhou, Abbeel, and Levine]{haarnoja2018soft}
Tuomas Haarnoja, Aurick Zhou, Pieter Abbeel, and Sergey Levine.
\newblock Soft actor-critic: Off-policy maximum entropy deep reinforcement learning with a stochastic actor.
\newblock In \emph{International conference on machine learning}, pages 1861--1870. PMLR, 2018.

\bibitem[He et~al.(2023)He, Bai, Xu, Yang, Zhang, Wang, Zhao, and Li]{he2023diffusion}
Haoran He, Chenjia Bai, Kang Xu, Zhuoran Yang, Weinan Zhang, Dong Wang, Bin Zhao, and Xuelong Li.
\newblock Diffusion model is an effective planner and data synthesizer for multi-task reinforcement learning.
\newblock \emph{arXiv preprint arXiv:2305.18459}, 2023.

\bibitem[Ho and Ermon(2016)]{ho2016generative}
Jonathan Ho and Stefano Ermon.
\newblock Generative adversarial imitation learning.
\newblock \emph{Advances in neural information processing systems}, 29, 2016.

\bibitem[Ho et~al.(2020)Ho, Jain, and Abbeel]{ho2020denoising}
Jonathan Ho, Ajay Jain, and Pieter Abbeel.
\newblock Denoising diffusion probabilistic models.
\newblock \emph{Advances in Neural Information Processing Systems}, 33:\penalty0 6840--6851, 2020.

\bibitem[Hong et~al.(2023)Hong, Dragan, and Levine]{hong2023offline}
Joey Hong, Anca Dragan, and Sergey Levine.
\newblock Offline rl with observation histories: Analyzing and improving sample complexity.
\newblock \emph{arXiv preprint arXiv:2310.20663}, 2023.

\bibitem[Hu et~al.(2021)Hu, Shen, Wallis, Allen-Zhu, Li, Wang, Wang, and Chen]{hu2021lora}
Edward~J Hu, Yelong Shen, Phillip Wallis, Zeyuan Allen-Zhu, Yuanzhi Li, Shean Wang, Lu~Wang, and Weizhu Chen.
\newblock Lora: Low-rank adaptation of large language models.
\newblock \emph{arXiv preprint arXiv:2106.09685}, 2021.

\bibitem[Huang et~al.(2024)Huang, Shen, Zhao, Yuan, and Tao]{huang2024solving}
Kaixin Huang, Li~Shen, Chen Zhao, Chun Yuan, and Dacheng Tao.
\newblock Solving continual offline reinforcement learning with decision transformer.
\newblock \emph{arXiv preprint arXiv:2401.08478}, 2024.

\bibitem[Janner et~al.(2021)Janner, Li, and Levine]{janner2021offline}
Michael Janner, Qiyang Li, and Sergey Levine.
\newblock Offline reinforcement learning as one big sequence modeling problem.
\newblock \emph{Advances in neural information processing systems}, 34:\penalty0 1273--1286, 2021.

\bibitem[Janner et~al.(2022)Janner, Du, Tenenbaum, and Levine]{janner2022planning}
Michael Janner, Yilun Du, Joshua~B Tenenbaum, and Sergey Levine.
\newblock Planning with diffusion for flexible behavior synthesis.
\newblock \emph{arXiv preprint arXiv:2205.09991}, 2022.

\bibitem[Kang et~al.(2023)Kang, Ma, Du, Pang, and Yan]{kang2023efficient}
Bingyi Kang, Xiao Ma, Chao Du, Tianyu Pang, and Shuicheng Yan.
\newblock Efficient diffusion policies for offline reinforcement learning.
\newblock \emph{arXiv preprint arXiv:2305.20081}, 2023.

\bibitem[Kaplanis et~al.(2019)Kaplanis, Shanahan, and Clopath]{kaplanis2019policy}
Christos Kaplanis, Murray Shanahan, and Claudia Clopath.
\newblock Policy consolidation for continual reinforcement learning.
\newblock \emph{arXiv preprint arXiv:1902.00255}, 2019.

\bibitem[Kaufmann et~al.(2023)Kaufmann, Bauersfeld, Loquercio, M{\"u}ller, Koltun, and Scaramuzza]{kaufmann2023champion}
Elia Kaufmann, Leonard Bauersfeld, Antonio Loquercio, Matthias M{\"u}ller, Vladlen Koltun, and Davide Scaramuzza.
\newblock Champion-level drone racing using deep reinforcement learning.
\newblock \emph{Nature}, 620\penalty0 (7976):\penalty0 982--987, 2023.

\bibitem[Kessler et~al.(2020)Kessler, Parker-Holder, Ball, Zohren, and Roberts]{kessler2020unclear}
Samuel Kessler, Jack Parker-Holder, Philip Ball, Stefan Zohren, and Stephen~J Roberts.
\newblock Unclear: A straightforward method for continual reinforcement learning.
\newblock In \emph{Proceedings of the 37th International Conference on Machine Learning}, 2020.

\bibitem[Kidambi et~al.(2020)Kidambi, Rajeswaran, Netrapalli, and Joachims]{kidambi2020morel}
Rahul Kidambi, Aravind Rajeswaran, Praneeth Netrapalli, and Thorsten Joachims.
\newblock Morel: Model-based offline reinforcement learning.
\newblock \emph{Advances in neural information processing systems}, 33:\penalty0 21810--21823, 2020.

\bibitem[Kirkpatrick et~al.(2017)Kirkpatrick, Pascanu, Rabinowitz, Veness, Desjardins, Rusu, Milan, Quan, Ramalho, Grabska-Barwinska, et~al.]{kirkpatrick2017overcoming}
James Kirkpatrick, Razvan Pascanu, Neil Rabinowitz, Joel Veness, Guillaume Desjardins, Andrei~A Rusu, Kieran Milan, John Quan, Tiago Ramalho, Agnieszka Grabska-Barwinska, et~al.
\newblock Overcoming catastrophic forgetting in neural networks.
\newblock \emph{Proceedings of the national academy of sciences}, 114\penalty0 (13):\penalty0 3521--3526, 2017.

\bibitem[Korycki and Krawczyk(2021)]{korycki2021class}
Lukasz Korycki and Bartosz Krawczyk.
\newblock Class-incremental experience replay for continual learning under concept drift.
\newblock In \emph{Proceedings of the IEEE/CVF Conference on Computer Vision and Pattern Recognition}, pages 3649--3658, 2021.

\bibitem[Kostrikov et~al.(2021)Kostrikov, Nair, and Levine]{kostrikov2021offline}
Ilya Kostrikov, Ashvin Nair, and Sergey Levine.
\newblock Offline reinforcement learning with implicit q-learning.
\newblock \emph{arXiv preprint arXiv:2110.06169}, 2021.

\bibitem[Kumar et~al.(2020)Kumar, Zhou, Tucker, and Levine]{kumar2020conservative}
Aviral Kumar, Aurick Zhou, George Tucker, and Sergey Levine.
\newblock Conservative q-learning for offline reinforcement learning.
\newblock \emph{Advances in Neural Information Processing Systems}, 33:\penalty0 1179--1191, 2020.

\bibitem[Laskin et~al.(2020)Laskin, Lee, Stooke, Pinto, Abbeel, and Srinivas]{laskin2020reinforcement}
Misha Laskin, Kimin Lee, Adam Stooke, Lerrel Pinto, Pieter Abbeel, and Aravind Srinivas.
\newblock Reinforcement learning with augmented data.
\newblock \emph{Advances in neural information processing systems}, 33:\penalty0 19884--19895, 2020.

\bibitem[Lee et~al.(2024)Lee, Cho, Kim, Gwak, Kim, Choo, Yun, and Yun]{lee2024plastic}
Hojoon Lee, Hanseul Cho, Hyunseung Kim, Daehoon Gwak, Joonkee Kim, Jaegul Choo, Se-Young Yun, and Chulhee Yun.
\newblock Plastic: Improving input and label plasticity for sample efficient reinforcement learning.
\newblock \emph{Advances in Neural Information Processing Systems}, 36, 2024.

\bibitem[Levine et~al.(2020)Levine, Kumar, Tucker, and Fu]{levine2020offline}
Sergey Levine, Aviral Kumar, George Tucker, and Justin Fu.
\newblock Offline reinforcement learning: Tutorial, review, and perspectives on open problems.
\newblock \emph{arXiv preprint arXiv:2005.01643}, 2020.

\bibitem[Li et~al.(2020)Li, Vuong, Liu, Liu, Ciosek, Christensen, and Su]{li2020multi}
Jiachen Li, Quan Vuong, Shuang Liu, Minghua Liu, Kamil Ciosek, Henrik Christensen, and Hao Su.
\newblock Multi-task batch reinforcement learning with metric learning.
\newblock \emph{Advances in Neural Information Processing Systems}, 33:\penalty0 6197--6210, 2020.

\bibitem[Li et~al.(2023)Li, Yu, Liang, He, Karampatziakis, Chen, and Zhao]{li2023loftq}
Yixiao Li, Yifan Yu, Chen Liang, Pengcheng He, Nikos Karampatziakis, Weizhu Chen, and Tuo Zhao.
\newblock Loftq: Lora-fine-tuning-aware quantization for large language models.
\newblock \emph{arXiv preprint arXiv:2310.08659}, 2023.

\bibitem[Liu et~al.(2023)Liu, Park, Azadi, Zhang, Chopikyan, Hu, Shi, Rohrbach, and Darrell]{liu2023more}
Xihui Liu, Dong~Huk Park, Samaneh Azadi, Gong Zhang, Arman Chopikyan, Yuxiao Hu, Humphrey Shi, Anna Rohrbach, and Trevor Darrell.
\newblock More control for free! image synthesis with semantic diffusion guidance.
\newblock In \emph{Proceedings of the IEEE/CVF Winter Conference on Applications of Computer Vision}, pages 289--299, 2023.

\bibitem[Mallya and Lazebnik(2018)]{mallya2018packnet}
Arun Mallya and Svetlana Lazebnik.
\newblock Packnet: Adding multiple tasks to a single network by iterative pruning.
\newblock In \emph{Proceedings of the IEEE conference on Computer Vision and Pattern Recognition}, pages 7765--7773, 2018.

\bibitem[Mangrulkar et~al.(2022)Mangrulkar, Gugger, Debut, Belkada, Paul, and Bossan]{peft}
Sourab Mangrulkar, Sylvain Gugger, Lysandre Debut, Younes Belkada, Sayak Paul, and Benjamin Bossan.
\newblock Peft: State-of-the-art parameter-efficient fine-tuning methods.
\newblock \url{https://github.com/huggingface/peft}, 2022.

\bibitem[Mendez and Eaton(2022)]{mendez2022reuse}
Jorge~A Mendez and Eric Eaton.
\newblock How to reuse and compose knowledge for a lifetime of tasks: A survey on continual learning and functional composition.
\newblock \emph{arXiv preprint arXiv:2207.07730}, 2022.

\bibitem[Meyer et~al.(2023)Meyer, White, and Machado]{meyer2023harnessing}
Edan Meyer, Adam White, and Marlos~C Machado.
\newblock Harnessing discrete representations for continual reinforcement learning.
\newblock \emph{arXiv preprint arXiv:2312.01203}, 2023.

\bibitem[Mnih et~al.(2015)Mnih, Kavukcuoglu, Silver, Rusu, Veness, Bellemare, Graves, Riedmiller, Fidjeland, Ostrovski, et~al.]{mnih2015human}
Volodymyr Mnih, Koray Kavukcuoglu, David Silver, Andrei~A Rusu, Joel Veness, Marc~G Bellemare, Alex Graves, Martin Riedmiller, Andreas~K Fidjeland, Georg Ostrovski, et~al.
\newblock Human-level control through deep reinforcement learning.
\newblock \emph{nature}, 518\penalty0 (7540):\penalty0 529--533, 2015.

\bibitem[Nguyen et~al.(2017)Nguyen, Li, Bui, and Turner]{nguyen2017variational}
Cuong~V Nguyen, Yingzhen Li, Thang~D Bui, and Richard~E Turner.
\newblock Variational continual learning.
\newblock \emph{arXiv preprint arXiv:1710.10628}, 2017.

\bibitem[Nguyen-Tang and Arora(2024)]{nguyen2024sample}
Thanh Nguyen-Tang and Raman Arora.
\newblock On sample-efficient offline reinforcement learning: Data diversity, posterior sampling, and beyond.
\newblock \emph{arXiv preprint arXiv:2401.03301}, 2024.

\bibitem[Ni et~al.(2023)Ni, Hao, Mu, Yuan, Zheng, Wang, and Liang]{ni2023metadiffuser}
Fei Ni, Jianye Hao, Yao Mu, Yifu Yuan, Yan Zheng, Bin Wang, and Zhixuan Liang.
\newblock Metadiffuser: Diffusion model as conditional planner for offline meta-rl.
\newblock \emph{arXiv preprint arXiv:2305.19923}, 2023.

\bibitem[Nichol and Dhariwal(2021)]{nichol2021improved}
Alexander~Quinn Nichol and Prafulla Dhariwal.
\newblock Improved denoising diffusion probabilistic models.
\newblock In \emph{International Conference on Machine Learning}, pages 8162--8171. PMLR, 2021.

\bibitem[Peng et~al.(2023)Peng, Giampouras, and Vidal]{peng2023ideal}
Liangzu Peng, Paris Giampouras, and Ren{\'e} Vidal.
\newblock The ideal continual learner: An agent that never forgets.
\newblock In \emph{International Conference on Machine Learning}, pages 27585--27610. PMLR, 2023.

\bibitem[Rafailov et~al.(2021)Rafailov, Yu, Rajeswaran, and Finn]{rafailov2021offline}
Rafael Rafailov, Tianhe Yu, Aravind Rajeswaran, and Chelsea Finn.
\newblock Offline reinforcement learning from images with latent space models.
\newblock In \emph{Learning for Dynamics and Control}, pages 1154--1168. PMLR, 2021.

\bibitem[Rakelly et~al.(2019)Rakelly, Zhou, Finn, Levine, and Quillen]{rakelly2019efficient}
Kate Rakelly, Aurick Zhou, Chelsea Finn, Sergey Levine, and Deirdre Quillen.
\newblock Efficient off-policy meta-reinforcement learning via probabilistic context variables.
\newblock In \emph{International conference on machine learning}, pages 5331--5340. PMLR, 2019.

\bibitem[Rigter et~al.(2022)Rigter, Lacerda, and Hawes]{rigter2022rambo}
Marc Rigter, Bruno Lacerda, and Nick Hawes.
\newblock Rambo-rl: Robust adversarial model-based offline reinforcement learning.
\newblock \emph{arXiv preprint arXiv:2204.12581}, 2022.

\bibitem[Rolnick et~al.(2019)Rolnick, Ahuja, Schwarz, Lillicrap, and Wayne]{rolnick2019experience}
David Rolnick, Arun Ahuja, Jonathan Schwarz, Timothy Lillicrap, and Gregory Wayne.
\newblock Experience replay for continual learning.
\newblock \emph{Advances in Neural Information Processing Systems}, 32, 2019.

\bibitem[Rombach et~al.(2022)Rombach, Blattmann, Lorenz, Esser, and Ommer]{rombach2022high}
Robin Rombach, Andreas Blattmann, Dominik Lorenz, Patrick Esser, and Bj{\"o}rn Ommer.
\newblock High-resolution image synthesis with latent diffusion models.
\newblock In \emph{Proceedings of the IEEE/CVF Conference on Computer Vision and Pattern Recognition}, pages 10684--10695, 2022.

\bibitem[Saharia et~al.()Saharia, Chan, Saxena, Li, Whang, Denton, Ghasemipour, Ayan, Mahdavi, Lopes, et~al.]{saharia2205photorealistic}
Chitwan Saharia, William Chan, Saurabh Saxena, Lala Li, Jay Whang, Emily Denton, Seyed Kamyar~Seyed Ghasemipour, Burcu~Karagol Ayan, S~Sara Mahdavi, Rapha~Gontijo Lopes, et~al.
\newblock Photorealistic text-to-image diffusion models with deep language understanding, 2022.
\newblock \emph{URL https://arxiv. org/abs/2205.11487}, 4.

\bibitem[Schwarz et~al.(2018)Schwarz, Czarnecki, Luketina, Grabska-Barwinska, Teh, Pascanu, and Hadsell]{schwarz2018progress}
Jonathan Schwarz, Wojciech Czarnecki, Jelena Luketina, Agnieszka Grabska-Barwinska, Yee~Whye Teh, Razvan Pascanu, and Raia Hadsell.
\newblock Progress \& compress: A scalable framework for continual learning.
\newblock In \emph{International conference on machine learning}, pages 4528--4537. PMLR, 2018.

\bibitem[Shin et~al.(2017)Shin, Lee, Kim, and Kim]{shin2017continual}
Hanul Shin, Jung~Kwon Lee, Jaehong Kim, and Jiwon Kim.
\newblock Continual learning with deep generative replay.
\newblock \emph{Advances in neural information processing systems}, 30, 2017.

\bibitem[Smith et~al.(2023{\natexlab{a}})Smith, Hsu, Zhang, Hua, Kira, Shen, and Jin]{smith2023continual}
James~Seale Smith, Yen-Chang Hsu, Lingyu Zhang, Ting Hua, Zsolt Kira, Yilin Shen, and Hongxia Jin.
\newblock Continual diffusion: Continual customization of text-to-image diffusion with c-lora.
\newblock \emph{arXiv preprint arXiv:2304.06027}, 2023{\natexlab{a}}.

\bibitem[Smith et~al.(2023{\natexlab{b}})Smith, Tian, Halbe, Hsu, and Kira]{smith2023closer}
James~Seale Smith, Junjiao Tian, Shaunak Halbe, Yen-Chang Hsu, and Zsolt Kira.
\newblock A closer look at rehearsal-free continual learning.
\newblock In \emph{Proceedings of the IEEE/CVF Conference on Computer Vision and Pattern Recognition}, pages 2409--2419, 2023{\natexlab{b}}.

\bibitem[Sohl-Dickstein et~al.(2015)Sohl-Dickstein, Weiss, Maheswaranathan, and Ganguli]{sohl2015deep}
Jascha Sohl-Dickstein, Eric Weiss, Niru Maheswaranathan, and Surya Ganguli.
\newblock Deep unsupervised learning using nonequilibrium thermodynamics.
\newblock In \emph{International Conference on Machine Learning}, pages 2256--2265. PMLR, 2015.

\bibitem[Song et~al.(2020)Song, Meng, and Ermon]{song2020denoising}
Jiaming Song, Chenlin Meng, and Stefano Ermon.
\newblock Denoising diffusion implicit models.
\newblock \emph{arXiv preprint arXiv:2010.02502}, 2020.

\bibitem[Sun et~al.(2023)Sun, Ma, Madaan, Bonatti, Huang, and Kapoor]{sun2023smart}
Yanchao Sun, Shuang Ma, Ratnesh Madaan, Rogerio Bonatti, Furong Huang, and Ashish Kapoor.
\newblock Smart: Self-supervised multi-task pretraining with control transformers.
\newblock \emph{arXiv preprint arXiv:2301.09816}, 2023.

\bibitem[Todorov et~al.(2012)Todorov, Erez, and Tassa]{todorov2012mujoco}
Emanuel Todorov, Tom Erez, and Yuval Tassa.
\newblock Mujoco: A physics engine for model-based control.
\newblock In \emph{2012 IEEE/RSJ international conference on intelligent robots and systems}, pages 5026--5033. IEEE, 2012.

\bibitem[Van~de Ven et~al.(2022)Van~de Ven, Tuytelaars, and Tolias]{van2022three}
Gido~M Van~de Ven, Tinne Tuytelaars, and Andreas~S Tolias.
\newblock Three types of incremental learning.
\newblock \emph{Nature Machine Intelligence}, 4\penalty0 (12):\penalty0 1185--1197, 2022.

\bibitem[Wang et~al.(2023{\natexlab{a}})Wang, Yang, Wen, Liu, and Qiao]{wang2023critic}
Yuanfu Wang, Chao Yang, Ying Wen, Yu~Liu, and Yu~Qiao.
\newblock Critic-guided decision transformer for offline reinforcement learning.
\newblock \emph{arXiv preprint arXiv:2312.13716}, 2023{\natexlab{a}}.

\bibitem[Wang et~al.(2022{\natexlab{a}})Wang, Hunt, and Zhou]{wang2022diffusion}
Zhendong Wang, Jonathan~J Hunt, and Mingyuan Zhou.
\newblock Diffusion policies as an expressive policy class for offline reinforcement learning.
\newblock \emph{arXiv preprint arXiv:2208.06193}, 2022{\natexlab{a}}.

\bibitem[Wang et~al.(2023{\natexlab{b}})Wang, Shen, Duan, Suo, Fang, Liu, and Gao]{wang2023distributionally}
Zhenyi Wang, Li~Shen, Tiehang Duan, Qiuling Suo, Le~Fang, Wei Liu, and Mingchen Gao.
\newblock Distributionally robust memory evolution with generalized divergence for continual learning.
\newblock \emph{IEEE Transactions on Pattern Analysis and Machine Intelligence}, 2023{\natexlab{b}}.

\bibitem[Wang et~al.(2023{\natexlab{c}})Wang, Yang, Shen, and Huang]{wang2023comprehensive}
Zhenyi Wang, Enneng Yang, Li~Shen, and Heng Huang.
\newblock A comprehensive survey of forgetting in deep learning beyond continual learning.
\newblock \emph{arXiv preprint arXiv:2307.09218}, 2023{\natexlab{c}}.

\bibitem[Wang et~al.(2022{\natexlab{b}})Wang, Chen, and Dong]{wang2022dirichlet}
Zhi Wang, Chunlin Chen, and Daoyi Dong.
\newblock A dirichlet process mixture of robust task models for scalable lifelong reinforcement learning.
\newblock \emph{IEEE Transactions on Cybernetics}, 2022{\natexlab{b}}.

\bibitem[Wei et~al.(2023)Wei, Sun, Zheng, Vemprala, Bonatti, Chen, Madaan, Ba, Kapoor, and Ma]{wei2023imitation}
Yao Wei, Yanchao Sun, Ruijie Zheng, Sai Vemprala, Rogerio Bonatti, Shuhang Chen, Ratnesh Madaan, Zhongjie Ba, Ashish Kapoor, and Shuang Ma.
\newblock Is imitation all you need? generalized decision-making with dual-phase training.
\newblock In \emph{Proceedings of the IEEE/CVF International Conference on Computer Vision}, pages 16221--16231, 2023.

\bibitem[Wo{\l}czyk et~al.(2021)Wo{\l}czyk, Zajac, Pascanu, Kucinski, and Mi{\l}o{\'s}]{wolczyk2021continual}
Maciej Wo{\l}czyk, Micha{\l} Zajac, Razvan Pascanu, {\L}ukasz Kucinski, and Piotr Mi{\l}o{\'s}.
\newblock Continual world: A robotic benchmark for continual reinforcement learning.
\newblock \emph{Advances in Neural Information Processing Systems}, 34:\penalty0 28496--28510, 2021.

\bibitem[Wu et~al.(2019)Wu, Tucker, and Nachum]{wu2019behavior}
Yifan Wu, George Tucker, and Ofir Nachum.
\newblock Behavior regularized offline reinforcement learning.
\newblock \emph{arXiv preprint arXiv:1911.11361}, 2019.

\bibitem[Xu et~al.(2022)Xu, Shen, Zhang, Lu, Zhao, Tenenbaum, and Gan]{xu2022prompting}
Mengdi Xu, Yikang Shen, Shun Zhang, Yuchen Lu, Ding Zhao, Joshua Tenenbaum, and Chuang Gan.
\newblock Prompting decision transformer for few-shot policy generalization.
\newblock In \emph{international conference on machine learning}, pages 24631--24645. PMLR, 2022.

\bibitem[Yang et~al.(2023)Yang, Zhou, Jiang, Long, and Shi]{yang2023continual}
Yijun Yang, Tianyi Zhou, Jing Jiang, Guodong Long, and Yuhui Shi.
\newblock Continual task allocation in meta-policy network via sparse prompting.
\newblock \emph{arXiv preprint arXiv:2305.18444}, 2023.

\bibitem[Yu et~al.(2020)Yu, Quillen, He, Julian, Hausman, Finn, and Levine]{yu2020meta}
Tianhe Yu, Deirdre Quillen, Zhanpeng He, Ryan Julian, Karol Hausman, Chelsea Finn, and Sergey Levine.
\newblock Meta-world: A benchmark and evaluation for multi-task and meta reinforcement learning.
\newblock In \emph{Conference on robot learning}, pages 1094--1100. PMLR, 2020.

\bibitem[Yue et~al.(2024)Yue, Liu, and Stone]{yue2024t}
William Yue, Bo~Liu, and Peter Stone.
\newblock t-dgr: A trajectory-based deep generative replay method for continual learning in decision making.
\newblock \emph{arXiv preprint arXiv:2401.02576}, 2024.

\bibitem[Zeng et~al.(2019)Zeng, Chen, Cui, and Yu]{zeng2019continual}
Guanxiong Zeng, Yang Chen, Bo~Cui, and Shan Yu.
\newblock Continual learning of context-dependent processing in neural networks.
\newblock \emph{Nature Machine Intelligence}, 1\penalty0 (8):\penalty0 364--372, 2019.

\bibitem[Zhang et~al.(2023{\natexlab{a}})Zhang, Zou, An, Liu, and Zhang]{zhang2023split}
Qizhe Zhang, Bocheng Zou, Ruichuan An, Jiaming Liu, and Shanghang Zhang.
\newblock Split \& merge: Unlocking the potential of visual adapters via sparse training.
\newblock \emph{arXiv preprint arXiv:2312.02923}, 2023{\natexlab{a}}.

\bibitem[Zhang et~al.(2022)Zhang, Wang, Liang, and Yuan]{zhang2022catastrophic}
Tiantian Zhang, Xueqian Wang, Bin Liang, and Bo~Yuan.
\newblock Catastrophic interference in reinforcement learning: A solution based on context division and knowledge distillation.
\newblock \emph{IEEE Transactions on Neural Networks and Learning Systems}, 2022.

\bibitem[Zhang et~al.(2023{\natexlab{b}})Zhang, Lin, Wang, Ye, Fu, Yang, Wang, Liang, Yuan, and Li]{zhang2023dynamics}
Tiantian Zhang, Zichuan Lin, Yuxing Wang, Deheng Ye, Qiang Fu, Wei Yang, Xueqian Wang, Bin Liang, Bo~Yuan, and Xiu Li.
\newblock Dynamics-adaptive continual reinforcement learning via progressive contextualization.
\newblock \emph{IEEE Transactions on Neural Networks and Learning Systems}, 2023{\natexlab{b}}.

\bibitem[Zhang et~al.(2023{\natexlab{c}})Zhang, Shen, Lin, Yuan, Wang, Li, and Ye]{zhang2023replay}
Tiantian Zhang, Kevin~Zehua Shen, Zichuan Lin, Bo~Yuan, Xueqian Wang, Xiu Li, and Deheng Ye.
\newblock Replay-enhanced continual reinforcement learning.
\newblock \emph{arXiv preprint arXiv:2311.11557}, 2023{\natexlab{c}}.

\bibitem[Zhu et~al.(2023)Zhu, Zhao, He, Zhong, Zhang, Yu, and Zhang]{zhu2023diffusion}
Zhengbang Zhu, Hanye Zhao, Haoran He, Yichao Zhong, Shenyu Zhang, Yong Yu, and Weinan Zhang.
\newblock Diffusion models for reinforcement learning: A survey.
\newblock \emph{arXiv preprint arXiv:2311.01223}, 2023.

\end{thebibliography}
\bibliographystyle{plainnat}

\clearpage

\section{Supplementary Material}

\subsection{Pseudocode of Continual Diffuser}\label{appendix of pseudocode}
The pseudocode for \ourmodel{} training is shown in Algorithm~\ref{ourmodel algorithm train}.
First of all, we process the datasets of $I$ tasks before training, including splitting the trajectories into equal sequences and normalizing the sequences to facilitate learning.
As shown in lines $9-24$, for each task $i$, we check the task index in the whole task sequence and sample different samples from the different buffers.
For example, for task $i,i>0$, we will perform experience rehearsal every $\upsilon$ train steps by sampling data from $\mathscr{D}_j, j\in{0, ..., i-1}$, where $j$ is sampled from $U(0, i-1)$.
Then, the networks $\epsilon_{\theta}$, $f_{task}(\phi)$, and $f_{time}(\varphi)$ are updated according to Equation~\eqref{single task DM loss} and Equation~\eqref{total loss func}.
After training on task $i$, we preserve a small portion ($\xi$) of the dataset of buffer $D_i$ as task $i$'s rehearsal buffer.
During the evaluation of multiple tasks (shown in Algorithm~\ref{ourmodel algorithm evaluation}), we successively generate decisions with \ourmodel{} and calculate the evaluation metrics.

\subsection{Implement Details}\label{Implementation Details}

\noindent{\textbf{Compute.}}~~~~Experiments are carried out on NVIDIA GeForce RTX 3090 GPUs and NVIDIA A10 GPUs.
Besides, the CPU type is Intel(R) Xeon(R) Gold 6230 CPU @ 2.10GHz. 
Each run of the experiments spanned about 24-72 hours, depending on the algorithm and the length of task sequences.

\noindent{\textbf{Hyperparameters.}}~~~~
In the implementation, we select the maximum diffusion steps as 200, and the default structure is Unet.
Then, in order to speed up the generation efficiency during evaluation, we consider the speed-up technique of DDIM~\cite{song2020denoising} and realize it in our method, thus accomplishing \textbf{19.043x} acceleration compared to the original generation method.
The sequence length is set to 48 in all experiments, where a larger sequence length can capture a more sophisticated distribution of trajectories and may also increase the computation burden.
We set the LoRA dimension as 64 for each module of down-sampling, middle, and up-sampling blocks, and the percent of LoRA parameters is approximately 12\% in our experiments.

\begin{table}[ht!]
\centering
\caption{The hyperparameters of \ourmodel{}.}
\label{our method hyper}
\begin{tabular}{l | l l}
\toprule
 & Hyperparameter & Value\\
\midrule
\multirow{8}{*}{Architecture}  & network backbone & Unet\\
                               & hidden dimension & 128 \\
                               & down-sampling blocks & 3 \\
                               & middle blocks & 2 \\
                               & up-sampling blocks & 2 \\
                               & convolution multiply & (1, 4, 8) \\
                               & normalizer & Gaussian normalizer \\
                               & sampling type of diffusion & DDIM \\      
\midrule
\multirow{13}{*}{Training} & condition guidance $\omega$ & 1.2 \\
 & max diffusion step $K$ & 200 \\
 & sequence length $T_e$ & 48 \\
 & loss function & MSE \\
 & learning rate & $3\cdot10^{-4}$ \\
 & batch size & 32 \\
 & optimizer & Adam \\
 & discount factor$\gamma$ & 0.99 \\
 & LoRA dimension & 64 \\
 & condition dropout & 0.25 \\
 & sampling speed-up stride & 10 \\
 & rehearsal frequency $\upsilon$ & 2 \\
 & rehearsal sample diversity $\xi$ & 0.1 \\
\bottomrule
\end{tabular}
\vspace{-0.4cm}
\end{table}

\begin{algorithm}
\caption{Training of Continual Diffuser (\ourmodel{})}
\label{ourmodel algorithm train}
\LinesNumbered
\KwIn{Noise prediction model $\epsilon_{\theta}$, task MLP $f_{task}(\phi)$, time MLP $f_{time}(\varphi)$, tasks set $\mathcal{M}_i, i\in\{1,...,I\}$, max diffusion step $K$, sequence length $T_e$, state dimension $d_s$, action dimension $d_a$, reply buffer $D_i, i\in\{1,...,I\}$, rehearsal frequency ($\upsilon$), rehearsal sample diversity ($\xi$), noise schedule $\alpha_{0:K}$ and $\beta_{0:K}$}
\KwOut{$\epsilon_{\theta}$, $f_{task}(\phi)$, $f_{time}(\varphi)$}
\textbf{Initialization:} $\theta$, $\phi$, $\varphi$ \\
// \textbf{Prepare for Training} \\
Separate the state-action trajectories of $D_i, i\in\{1,...,I\}$ into state-action sequences with length $T_e$ \\
Normalize state-action sequences to obey Gaussian distribution \\ 
// \textbf{Training} \\
\For{each task $i$}
{
   \For{each train epoch}
   {
       \For{each train step $m$}
       {
           \eIf{$i>0$ and $m$ \% $\upsilon$ == 0}
           {
               Sample j from $\{0, ..., i-1\}$ \\
               Sample $b$ sequences $\tau_j^{0}=\{s_{j,t:t+T_e}, a_{j,t:t+T_e}\}\in\mathbb{R}^{b\times T_e\times (d_s+d_a)}$ from task $j$'s rehearsal buffer $\mathscr{D}_j, j<i$ \\
           }
           {
            Sample $b$ sequences $\tau_i^{0}=\{s_{i,t:t+T_e}, a_{i,t:t+T_e}\}\in\mathbb{R}^{b\times T_e\times (d_s+d_a)}$ from task $i$'s buffer $D_i$ \\
            }
           Obtain the corresponding task conditions $\mathcal{C}_{task}$ \\
           Sample diffusion time step $k\sim \text{Uniform}(K)$ \\
           Obtain $\tau_i^{k}$ or $\tau_j^{k}$ by adding noise to $\tau_i^{0}$ or $\tau_j^{0}$ \\
           Sample Gaussian noise $\epsilon \sim \mathcal{N}(0,\bm{I}), \epsilon\in\mathbb{R}^{b\times T_e\times (d_s+d_a)}$ \\
           Train $\epsilon_{\theta}$, $f_{task}(\phi)$, and $f_{time}(\varphi)$ according to Equation~\eqref{single task DM loss} and Equation~\eqref{total loss func} \\ 
       }
       Save model periodically \\
   }
   Preserve a small portion ($\xi$) of the dataset of buffer $D_i$ as task $i$'s rehearsal buffer $\mathscr{D}_i$ \\
}
\end{algorithm}

\begin{algorithm}
\caption{Evaluation of Continual Diffuser (\ourmodel{})}
\label{ourmodel algorithm evaluation}
\LinesNumbered
\KwIn{Well trained noise prediction model $\epsilon_{\theta}$, task MLP $f_{task}(\phi)$, time MLP $f_{time}(\varphi)$, tasks set $\mathcal{M}_i, i\in\{1,...,I\}$, max diffusion step $K$, noise schedule $\alpha_{0:K}$ and $\beta_{0:K}$}
// \textbf{Prepare for Evaluation} \\
Normalize state-action sequences to obey Gaussian distribution \\ 
// \textbf{Evaluation} \\
\For{each evaluation task $i$}
{
    \For{each evaluation step}
    {
        Receive state $s_{i,t}$ and task identify from the task $i$ \\
        Obtain the corresponding task conditions $\mathcal{C}_{task}$ \\
        Let $k=K$ \\ 
        Sample $\hat{\tau}^{k}_i\in\mathbb{R}^{1\times T_e\times (d_s+d_a)}$ from normal distribution $\mathcal{N}(0, \bm{I})$ \\ 
        Replace the first state of $\hat{\tau}^{k}_i$ with $s_{i,t}$ \\
        \For{each generation step $k$}
        {
            Generate sequences $\hat{\tau}^{k-1}_i$ with $\epsilon_{\theta}$, $f_{task}(\phi)$, and $f_{time}(\varphi)$ according to Equation~\eqref{generation equation} \\
            Replace the first state of $\hat{\tau}^{k-1}_i$ with $s_{i,t}$ \\
        }
        Perform the first action of $\hat{\tau}^{k-1}_i$ in the task $i$ \\
        Observe reward $r$ from the task $i$ \\
    }
    Record the success rate on task $i$ \\
}
Calculate the total mean success rate \\
\end{algorithm}

\subsection{Related Work}

\noindent\textbf{Diffusion-Based Models for RL.}~~~
Diffusion models have made big progress in many fields, such as image synthesis and text generation~\cite{ho2020denoising, saharia2205photorealistic, nichol2021improved, beeson2023balancing, sohl2015deep, rombach2022high}.
Recently, a series of works have demonstrated the tremendous potential of diffusion-based models in offline RL tasks such as goal-based planning, composable constraint combination, scalable trajectory generation, and complex skill synthesis~\cite{janner2022planning, fontanesi2019reinforcement, ajay2022conditional, chi2023diffusion}.
For example, \citet{janner2022planning} propose to use the value function as the guide during trajectory generation, effectively reducing the effects of out-of-distribution actions and reaching remarkable performance in offline RL tasks.
% \citet{ajay2022conditional} show the composability of constraints and skills during training diffusion models and achieve better performance in robotics manipulation. 
Besides, diffusion models can also be used as policies to model the multimodal distribution from states to actions and as planners to perform long-horizon planning~\cite{kang2023efficient, wang2022diffusion, he2023diffusion, ni2023metadiffuser}.
For instance, \citet{kang2023efficient} use diffusion models as policies to model the distribution from states to actions, while \citet{he2023diffusion} endow diffusion models with the ability to perform planning and data augmentation with different task-specific prompts.

\noindent\textbf{Continual Learning in RL.}~~~
Continual learning (CL) aims to solve multi-tasks that come sequentially with explicit boundaries (task-aware CL) or implicit boundaries (task-free CL) and achieve no catastrophic forgetting and good task transferring (i.e., plasticity-stability dilemma) at the same time~\cite{zhang2023replay, meyer2023harnessing, wang2023distributionally}.
Multitask learning methods~\cite{he2023diffusion, laskin2020reinforcement} are usually regarded as the upper bound of continual learning.
Existing studies for continual RL can be roughly classified into three categories: Structure-based methods focus on novel model structures such as sub-networks, mixture-of-experts, hypernetworks, and low-rank adaptation~\cite{wang2022dirichlet, zhang2023split, smith2023continual, mallya2018packnet}. 
Regularization-based methods propose using auxiliary regularization loss to constrain the policy optimization and avoid catastrophic forgetting during training~\cite{zhang2023dynamics, zhang2022catastrophic, kessler2020unclear, kaplanis2019policy}. 
Rehearsal-based methods preserve experiences of previous tasks or train generative models that can produce pseudo-samples to maintain knowledge of past tasks~\cite{korycki2021class, smith2023closer, atkinson2021pseudo, peng2023ideal}.
Besides, recent plasticity-preserving studies~\cite{lee2024plastic, foret2020sharpness} reveal that the plasticity of models can be enhanced by weight re-initialization and noisification when facing the early interactions overfitting within a single task.

\noindent\textbf{Offline RL.}~~~
Offline RL mainly focuses on how to train optimal policies with previously collected large datasets without expensive and risky data collection processes~\cite{wang2023critic, levine2020offline, kostrikov2021offline, kumar2020conservative, ghosh2022offline}. 
It, however, remains a huge challenge for training when facing the distribution shift between the learned policy and the data-collected policy and the overestimation of out-of-distribution (OOD) actions~\cite{hong2023offline, kostrikov2021offline}.
To solve these issues, previous studies on offline-RL tasks generally rely on methods from constrained optimization, safe learning, imitation learning, and amendatory estimation~\cite{kostrikov2021offline, kumar2020conservative, wu2019behavior, ghosh2022offline}.  
Besides, planning and optimizing in the world model with limited interactions also serves as a promising way to train satisfactory policies~\cite{rigter2022rambo, kidambi2020morel, rafailov2021offline}.
Recently, sequential modeling has been proposed to fit the joint state-action distribution over the trajectories with transformer-based models and diffusion-based models~\cite{janner2021offline, nguyen2024sample, chen2021decision, ajay2022conditional, zhu2023diffusion, janner2022planning}.

\subsection{Additional Experiments}\label{Additional Experiments}

\begin{figure*}[t!]
% \hspace{-1em}
%\vspace{-0.5em}
 \begin{center}
 \includegraphics[angle=0,width=0.99\textwidth]{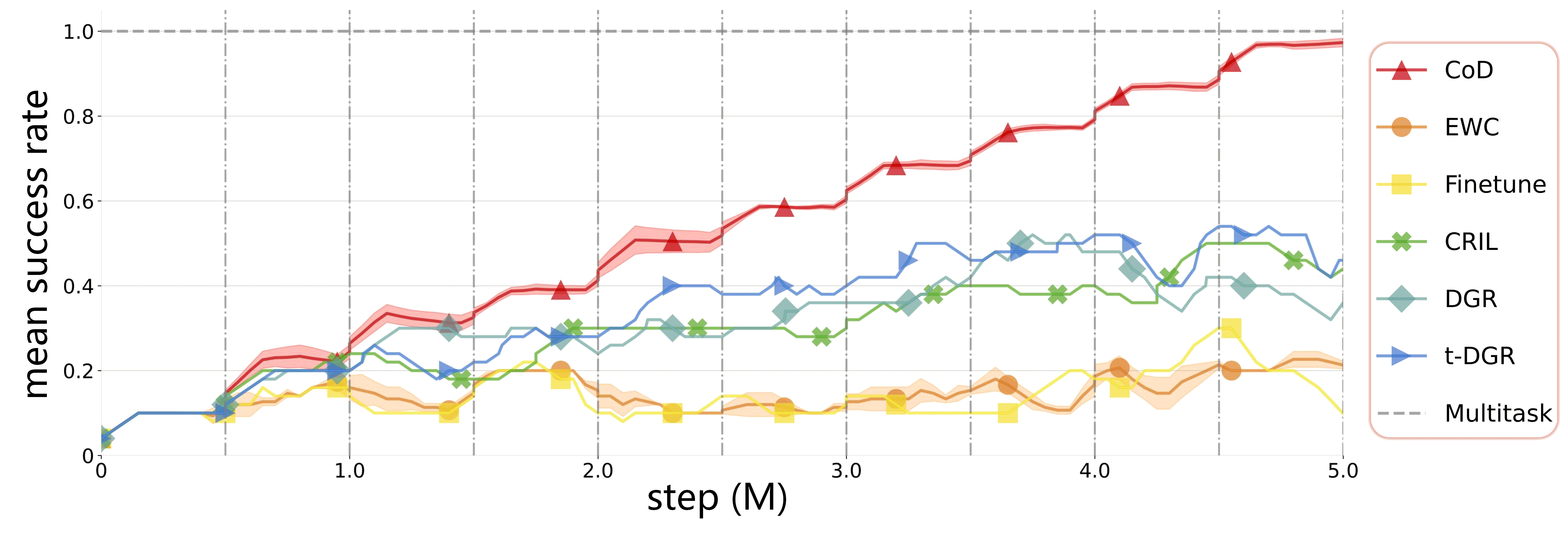}
 \caption{The comparison of our method \ourmodel{} and other baselines on CW10 where these baselines are trained with offline datasets and are trained with 500k gradient steps on each task.}
 \label{cw10 total success rate comparison of offline}
 \end{center}
 % \vspace{-0.3cm}
 % \vspace{-0.5em}
 \end{figure*}

\begin{figure*}[t!]
% \hspace{-1em}
%\vspace{-0.5em}
 \begin{center}
 \includegraphics[angle=0,width=0.99\textwidth]{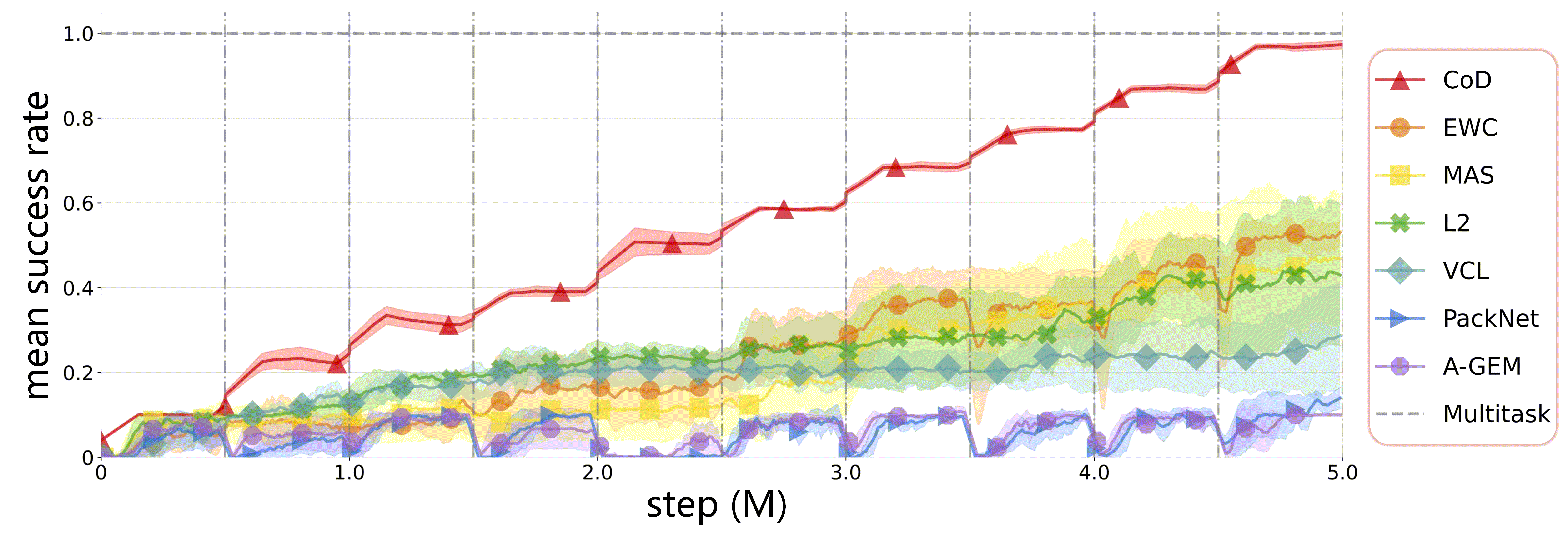}
 \caption{The comparison of our method \ourmodel{} and other baselines on CW10 where these baselines are trained with online environments and are trained with 500k interaction steps on each task.}
 \label{cw10 total success rate comparison of online}
 \end{center}
 % \vspace{-0.3cm}
 % \vspace{-0.5em}
 \end{figure*}

\noindent{\textbf{Offline Continual World Results on CW10.}}~~~~
We report the performance on CW10 in Figure~\ref{cw10 total success rate comparison of offline} when the baselines are trained with offline datasets.
The results show that the learning speed of our method (\ourmodel{}) is much more efficient than other baselines when executing the same gradient updates.
Besides, we can observe that the performance of generative methods is more effective than non-generative methods, which shows the powerful expressiveness of generative models in modeling complex environments and generating pseudo-samples with high fidelity.

\noindent{\textbf{Online Continual World Results on CW10.}}~~~~
Apart from the offline comparison, we also modified the original baselines and conducted experiments on CW10, where several new online baselines were introduced.
Similarly, the results in Figure~\ref{cw10 total success rate comparison of online} also show that our method (\ourmodel{}) surpasses the baselines by a large margin, illustrating the superiority of \ourmodel{}.
We do not incorporate several offline baselines trained with generative models into online comparison because the generative process consumes much more time for interaction, which exceeds the tolerable range of training.
These baselines trained with generative models are more suitable for training on offline datasets.

\noindent{\textbf{Mixed Dataset Training Analysis.}}~~~~
We can classify the training under the sub-optimal demonstrations into two situations.
You can click here to return to Section~\ref{Results} quickly for continual reading of the main body.

The first is learning from noise datasets. In order to simulate the training under the sub-optimal demonstrations, we insert noise into the observations of the current dataset to obtain sub-optimal demonstrations, i.e., $\bar{o} = o + clip(\eta * \mathcal{N}(0, I), \rho)$. The larger noise denotes datasets with lower quality. We report the results in Table~\ref{mied dataset training with noise}. The results illustrate that the performance decreases with the noise increasing, which inspires us to find additional techniques to reduce the influence of the noise on samples, such as adding an extra denoising module before diffuser training.

\begin{table*}[h!]
\centering
% \vspace{-1em}
\small
\caption{The experiments of \ourmodel{} when training with noise datasets on the Ant-dir tasks.}
\label{mied dataset training with noise}
\resizebox{0.6\textwidth}{!}{
\begin{tabular}{l | r | r | r}
\toprule
\specialrule{0em}{1.5pt}{1.5pt}
\toprule
Noise level $\eta$ & 0 & 0.1 & 0.5 \\
Bound $\rho$ & - & (-0.5, 0.5) & (-1.0, 1.0) \\
Score & 478.19\tiny{$\pm$15.84} & 247.41\tiny{$\pm$5.48} & 163.00\tiny{$\pm$5.16} \\
\bottomrule
\specialrule{0em}{1.5pt}{1.5pt}
\bottomrule
\end{tabular}}
% \vspace{-0.2cm}
\end{table*}

The second is learning from datasets sampled with mixed-quality policies. 
We construct the `medium' datasets on several Continual World tasks (CW4) to show the performance on the mixed-quality datasets, where the trajectories come from a series of behavior policies during the training stage. 
With the training stage going, we update the policy network many times, and each gradient update step will be regarded as generating a new behavior policy. Then, the performance of the policy will be improved. 
Next, we use the behavior policies whose performance ranges from medium to well-trained performance to collect `medium' datasets, i.e., the `medium' datasets contain unsuccessful trajectories and successful trajectories simultaneously (Refer to Table~\ref{The information statistics of offline continual world datasets} for more statistics.).
Based on the mixed-quality CW4 datasets, we adopt IL as the baseline and compare our method with IL. 
The corresponding experimental results are shown in Table~\ref{The performance comparison on offline and online} (c). 
The results show that our method (\ourmodel{}) can achieve better performance than the baseline in the `medium' dataset quality setting, which shows its effectiveness.

\noindent{\textbf{Plasticity Comparison.}}~~~~
In order to compare the plasticity of our method and representative plasticity-preserving methods~\cite{lee2024plastic, foret2020sharpness}, we conduct the experiments on the Ant-dir environment with task setting as `10-15-19-25', which is the same as the setting in the main body.
The results are reported in Table~\ref{plasticity comparison}, where the final performance means evaluation on all tasks after the whole training on all tasks and the performance gain of plasticity (task-level) is calculated according to mean(P(train15test15) - P(train10test15) + P(train19test19) - P(train15test19) + P(train25test25) - P(train19test25)). 
The results illustrate that our model reaches better final performance than PLASTIC and SAM. Besides, in the task-level plasticity performance comparison, our method also obtains a higher score. 
Although PLASTIC and SAM do not perform well here, it's worth noting that PLASTIC and SAM are not designed to resolve continual learning under changing tasks but to address early interactions overfitting within a single task. 
The granularity of plasticity referred to in \ourmodel{} is larger than that in PLASTIC and SAM.
Click here to return to Section~\ref{Results} quickly for continual reading of the main body.

\begin{table*}[h!]
\centering
% \vspace{-1em}
\small
\caption{The comparison of our method and plasticity-preserving methods PLASTIC and SAM on the Ant-dir environment. We report the performance of PLASTIC and SAM with online and offline training under the continual learning setting.}
\label{plasticity comparison}
\resizebox{0.99\textwidth}{!}{
\begin{tabular}{l | r  r  r  r  r  r}
\toprule
\specialrule{0em}{1.5pt}{1.5pt}
\toprule
Model & \ourmodel{} & \makecell[r]{Diffuser-w/o \\rehearsal} & \makecell[r]{PLASTIC \\(online)} & \makecell[r]{SAM \\(online)} & \makecell[r]{PLASTIC \\(offline)} & \makecell[r]{SAM \\(offline)} \\
\midrule
Final performance & 478.19\tiny{$\pm$15.84} & 270.44\tiny{$\pm$5.54} & 201.45\tiny{$\pm$0.56} & 202.17\tiny{$\pm$0.46} & 186.71\tiny{$\pm$4.55} & 187.19\tiny{$\pm$4.53} \\
\midrule
\makecell[l]{Performance gain of \\plasticity (task-level)} & 407.84 & 348.15 & 8.70 & 10.94 & 4.60 & 55.77 \\
\bottomrule
\specialrule{0em}{1.5pt}{1.5pt}
\bottomrule
\end{tabular}}
% \vspace{-0.2cm}
\end{table*}

\noindent{\textbf{Parameters Sensitivity on Ant-dir.}}~~~~
In Section~\ref{Ablation Study}, we conduct the parameter sensitivity analysis on CW to show the effects of rehearsal frequency $\upsilon$ and rehearsal diversity $\xi$.
We also report the results of parameters sensitivity on Ant-dir in Figure~\ref{parameters sensitivity of Ant-dir} and Table~\ref{absolute performance of parameters sensitivity of Ant-dir}, where $\upsilon=inf$ means we do not perform rehearsal during training.
The results show that with the increase of $\upsilon$, the performance declines because the model can not use previous datasets to strengthen its memory in time.

\begin{table*}[h!]
    \vspace{-0.3cm}
    \centering
    \small
    \caption{The ablation study of \ourmodel{}.}
    \label{ablation study}
    \resizebox{0.5\textwidth}{!}{
    \begin{tabular}{l | r | r | r}
    \toprule
    \specialrule{0em}{1.5pt}{1.5pt}
    \toprule
    Method & \multicolumn{3}{c}{Mean episode return}\\
    \midrule
    Task & Ant-dir & CW10 & CW20 \\
    \midrule[1pt]
    \makecell[l]{CoD-w/o \\rehearsal} & 270.44\tiny{$\pm$5.54} & 0.20\tiny{$\pm$0.01} & 0.18\tiny{$\pm$0.01} \\
    \ourmodel{} (\textbf{Ours}) & 478.19\tiny{$\pm$15.84} & 0.98\tiny{$\pm$0.01} & 0.98\tiny{$\pm$0.01} \\
    \bottomrule
    \specialrule{0em}{1.5pt}{1.5pt}
    \bottomrule
    \end{tabular}}
    % \vspace{-0.3cm}
\end{table*}

\begin{table*}[t!]
\centering
% \vspace{-1em}
\small
\caption{The absolute performance of parameters sensitivity of Ant-dir.}
\label{absolute performance of parameters sensitivity of Ant-dir}
\resizebox{0.7\textwidth}{!}{
\begin{tabular}{l | l | r | r | r | r}
\toprule
\specialrule{0em}{1.5pt}{1.5pt}
\toprule
\multicolumn{2}{c|}{}& \multicolumn{4}{c}{$\xi$}\\
\midrule
\multicolumn{2}{c|}{}& 1\% & 5\% & 10\% & 20\% \\
\midrule
 \multirow{4}{*}{$\upsilon$} & 2 & 383.07\tiny{$\pm$8.71} & 465.82\tiny{$\pm$5.03} & 475.82\tiny{$\pm$8.14} & 493.83\tiny{$\pm$3.17}  \\
   & 10 & 344.73\tiny{$\pm$0.00} & 381.27\tiny{$\pm$0.00} & 368.42\tiny{$\pm$0.00} & 369.94\tiny{$\pm$0.00}   \\
   & 14 & 331.29\tiny{$\pm$0.00} & 351.79\tiny{$\pm$0.00} & 345.71\tiny{$\pm$0.00} & 352.32\tiny{$\pm$0.00}   \\
   & inf & 271.79\tiny{$\pm$8.48} & 271.79\tiny{$\pm$8.48} & 271.79\tiny{$\pm$8.48} & 271.79\tiny{$\pm$8.48}   \\
\bottomrule
\specialrule{0em}{1.5pt}{1.5pt}
\bottomrule
\end{tabular}}
% \vspace{-0.2cm}
\end{table*}

\noindent{\textbf{Efficiency Analysis of Generation Speed.}}~~~~
The generation process of diffusion models is indeed computationally intensive because the mechanism of generation requires multiple rounds to generate a sequence. However, we can draw inspiration from previous studies \cite{nichol2021improved, song2020denoising} in related domains and accelerate the generation process. For example, we can reduce the reverse diffusion step from 200 to 10. To show the efficiency of accelerating during the generation process, we conduct a comparison of generation speed. We report the results in Table~\ref{Efficiency Analysis of Generation Speed}, where the 200 diffusion steps setting is the original version, and the 10 diffusion steps setting is our accelerated version. In the experiments of our manuscript, we adopt the 10 diffusion steps setting, which improves the sampling speed (\textbf{19.043×}) with a larger margin than the original sampling version. It's worth noting that our implemented accelerate technique can also use other diffusion steps settings, but we find that 10 diffusion steps setting performs well on performance and generation efficiency.

\begin{table*}[h!]
\centering
% \vspace{-1em}
\small
\caption{The comparison of generation speed with different generation steps. In the main body of our manuscript, we use the 10 diffusion steps setting for all experiments.}
\label{Efficiency Analysis of Generation Speed}
\resizebox{0.99\textwidth}{!}{
\begin{tabular}{l | r  r  r  r  r}
\toprule
\specialrule{0em}{1.5pt}{1.5pt}
\toprule
Diffusion steps & 200 (original) & 100 & 50 & 25 & 10 \\
\midrule
\makecell[l]{Time consumption of \\ per generation (s)} & 3.085\tiny{$\pm$0.077} & 1.839\tiny{$\pm$0.116} & 0.850\tiny{$\pm$0.007} & 0.394\tiny{$\pm$0.006} & 0.159\tiny{$\pm$0.005} \\
\midrule
Speed-up ratio & 1× & 1.678× & 3.629× & 7.830× & 19.043× \\
\bottomrule
\specialrule{0em}{1.5pt}{1.5pt}
\bottomrule
\end{tabular}}
% \vspace{-0.2cm}
\end{table*}

\noindent{\textbf{Ablation Study on Mixed Datasets.}}~~~~
In Table~\ref{Ablation study on the mixed datasets}, we report the effects of rehearsal sample diversity $\xi$ on the `medium' datasets. From the results, we can see that increasing the rehearsal sample diversity is beneficial to the performance, which is in line with the experiments in the main body of our manuscript. Besides, the results also show that our method (\ourmodel{}) can reach a better plasticity-stability trade-off than the baseline in the `medium' dataset quality setting.

\begin{table*}[h!]
\centering
% \vspace{-1em}
\small
\caption{The ablation study of \ourmodel{} on `medium' CW4 datasets. We select the continual tasks setting as CW4 ({``hammer-v1'', ``push-wall-v1'', ``faucet-close-v1'', ``push-back-v1''}), where the `medium' experiences come from the behavior policies from the middle training stage to the end training stage.}
\label{Ablation study on the mixed datasets}
\resizebox{0.7\textwidth}{!}{
\begin{tabular}{l | r | r | r | r | r | r}
\toprule
\specialrule{0em}{1.5pt}{1.5pt}
\toprule
Model & $\upsilon$ & $\xi$ & P $\uparrow$ & FT $\uparrow$ & F $\downarrow$ & P+FT-F $\uparrow$ \\
\midrule
CoD & 2 & 1\% & 0.85\tiny{$\pm$0.02} & 0.60\tiny{$\pm$0.13} & 0.05\tiny{$\pm$0.01} & 1.40 \\
CoD & 2 & 10\% & 0.90\tiny{$\pm$0.02} & 0.60\tiny{$\pm$0.12} & -0.01\tiny{$\pm$0.01} & 1.51 \\
IL & 2 & 1\% & 0.57\tiny{$\pm$0.19} & 0.12\tiny{$\pm$0.54} & 0.18\tiny{$\pm$0.09} & 0.51 \\
IL & 2 & 10\% & 0.63\tiny{$\pm$0.17} & 0.28\tiny{$\pm$0.27} & 0.28\tiny{$\pm$0.18} & 0.63 \\
\bottomrule
\specialrule{0em}{1.5pt}{1.5pt}
\bottomrule
\end{tabular}}
% \vspace{-0.2cm}
\end{table*}

\noindent{\textbf{Computation Costs Analysis.}}~~~~
In order to show the consumption of computational costs, we report the comparison of computation costs during the training stage in Table~\ref{Computation Costs Analysis}, where we obtain the statistical data with `wandb.'
The results show that increasing the rehearsal samples does not significantly increase computation costs and training time.

\begin{table*}[h!]
\centering
% \vspace{-1em}
\small
\caption{The comparison of computation costs and training time using our method with different hyperparameter settings. Experiments are carried out on NVIDIA GeForce RTX 3090 GPUs and NVIDIA A10 GPUs. Besides, the CPU type is Intel(R) Xeon(R) Gold 5220 CPU @ 2.20GHz. We report the results according to `wand.'}
\label{Computation Costs Analysis}
\resizebox{0.99\textwidth}{!}{
\begin{tabular}{l | r  r  r  r  r  r  r  r}
\toprule
\specialrule{0em}{1.5pt}{1.5pt}
\toprule
$\upsilon$ & 2 & 2 & 2 & 2 & 14 & 10 & 6 & inf\\
\midrule
$\xi$ & 20 & 10 & 5 & 1 & 10 & 10 & 10 & - \\
\midrule
\makecell[c]{Process memory \\ in use (non-swap) \\ (MB) }& 19914.52 & 20058.77 & 20023.48 & 20045.32 & 20010.42 & 20026.34 & 20049.25 & 19983.4 \\
\midrule
Train time (h) & 143.713 & 144.488 & 143.681 & 143.398 & 145.081 & 145.973 & 146.397 & 152.274 \\
\bottomrule
\specialrule{0em}{1.5pt}{1.5pt}
\bottomrule
\end{tabular}}
% \vspace{-0.2cm}
\end{table*}

\subsection{Statistics of Continual Offline RL Benchmarks}\label{Statistics of Continual Offline RL Benchmarks}

\begin{table}[h]
\centering
% \vspace{-0.3em}
\small
\caption{The total statistics of our benchmark.}
\label{total statistics of the benchmark}
\resizebox{0.6\textwidth}{!}{
\begin{tabular}{l | r | r | r}
\toprule
\specialrule{0em}{1.5pt}{1.5pt}
\toprule
Environment & Tasks number & Quality & \makecell[c]{Samples per task}\\
\midrule[1pt]
\multirow{2}{*}{\makecell[c]{Continual World}} & 88 & expert & 1M\\
 & 4 & medium & 0.4M\\
 \midrule
Ant-dir & 40 & expert & 0.2M\\
Cheetah-dir & 2 & expert & 0.2M\\
\bottomrule
\specialrule{0em}{1.5pt}{1.5pt}
\bottomrule
\end{tabular}}
% \vspace{-0.2cm}
\end{table}

To take advantage of the potential of diffusion models, we first collect an offline benchmark that contains dozens of tasks from multiple domains, such as Continual World and Gym-MuJoCo~\cite{wolczyk2021continual, todorov2012mujoco}.
In order to collect the interaction data, we trained Soft Actor-Critic on each task for approximately 1M time steps~\cite{haarnoja2018soft}.
Totally, the benchmark contains 90 tasks, where 88 tasks come from Continual World, 2 tasks come from Gym-MuJoCo.

\begin{figure*}[h]
% \hspace{-1em}
%\vspace{-0.5em}
 \begin{center}
 \includegraphics[angle=0,width=0.95\textwidth]{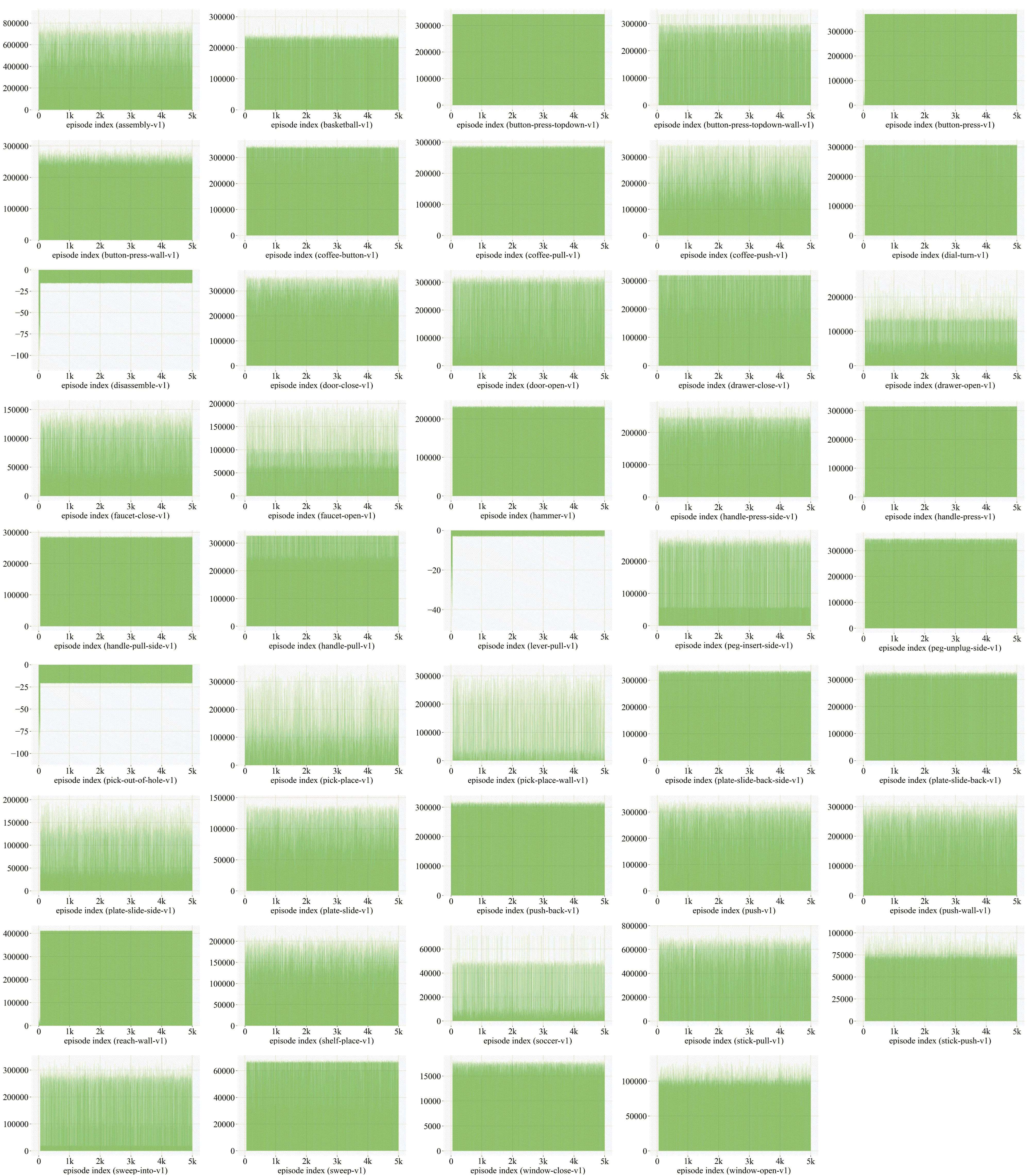}
 \caption{The return statistics of the Continual World-v1. We calculate the episode return of the Continual World datasets and report the corresponding histogram.}
 \label{continual world offline dataset return analysis}
 \end{center}
 \vspace{-0.3cm}
 % \vspace{-0.5em}
 \end{figure*}

\begin{figure*}[h]
% \hspace{-1em}
%\vspace{-0.5em}
 \begin{center}
 \includegraphics[angle=0,width=0.95\textwidth]{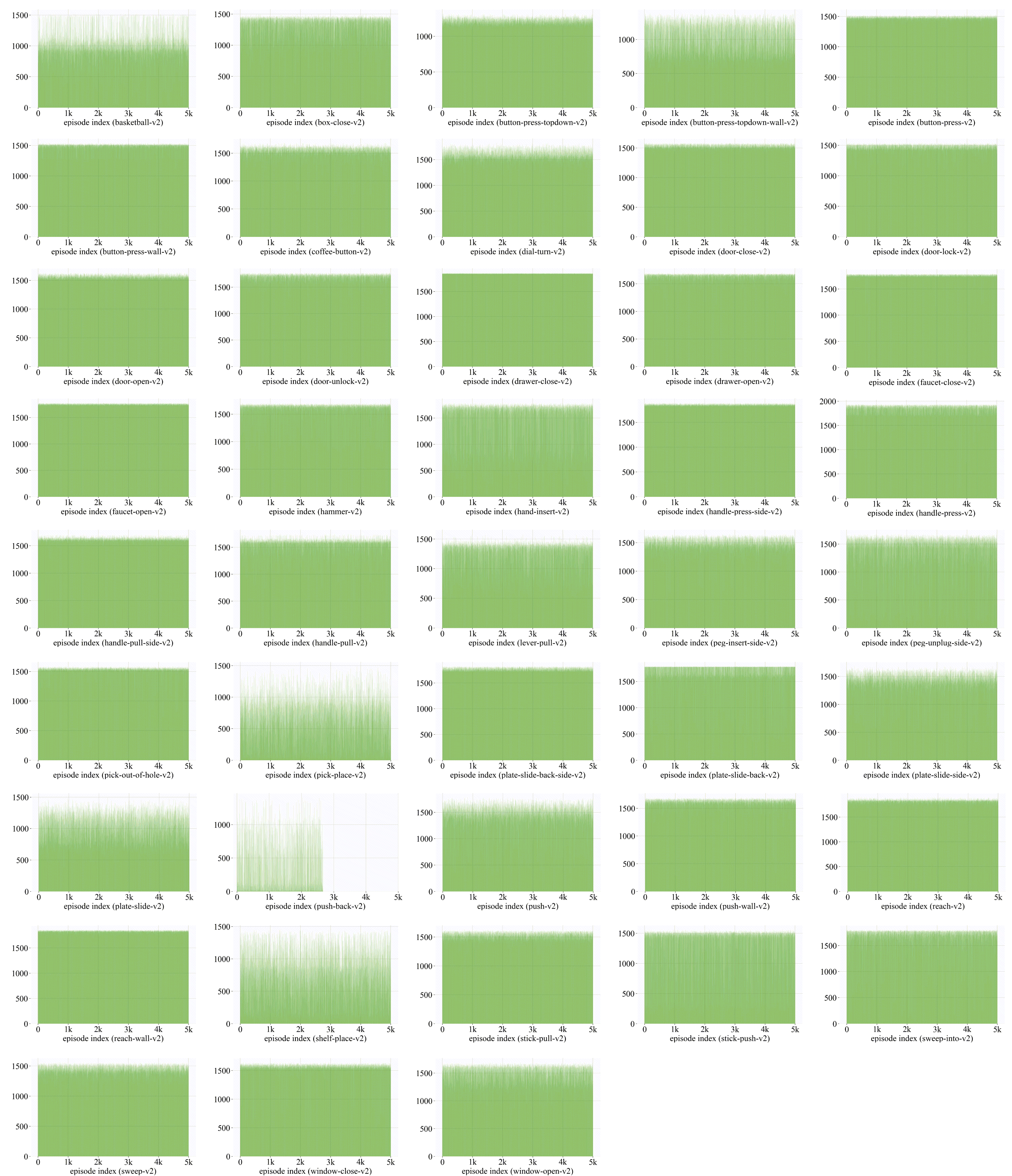}
 \caption{The return statistics of the Continual World-v2. We calculate the episode return of the Continual World datasets and report the corresponding histogram.}
 \label{continual world-v2 offline dataset return analysis}
 \end{center}
 \vspace{-0.3cm}
 % \vspace{-0.5em}
 \end{figure*}

Specifically, CW~\cite{wolczyk2021continual} tasks are constructed based on Meta-World~\cite{yu2020meta}.
CW consists of many realistic robotic manipulation 
tasks such as Pushing, Reaching, Door Opening, Pick, and Place.
CW is convenient for training and evaluating the abilities of forward transfer and forgetting because the state and action space are the same across all the tasks.
In our benchmark, we collect ``expert'' and ``medium'' datasets, where the episodic time limit is set to 200, and the evaluation time step is set to 1M and 0.4M for ``expert" and ``medium'' datasets, respectively.
Thus, we obtain 5000 and 2000 episodes for these two quality tasks, as shown in Table~\ref{The information statistics of offline continual world datasets} and Table~\ref{The information statistics of offline continual world v2 datasets}, in which we also report the mean success rate of these two qualities dataset.
Besides, we also provide the return information of all datasets in Figure~\ref{continual world offline dataset return analysis} and Figure~\ref{continual world-v2 offline dataset return analysis}.
Out of these tasks defined in Meta-World, we usually select ten tasks from them as the setting of continual learning, i.e., CW10, and CW20 denotes the setting of two CW10.

Aligning with the traditional definition of various CL settings~\cite{van2022three}, this benchmark supports constructing task-incremental CORL (TICORL), domain-incremental CORL (DICORL), and class-incremental CORL (CICORL) settings. 
Researchers can use these datasets in any sequence or length for CL tasks to test the plasticity-stability trade-off of their proposed methods. 
The future expansion plan of this benchmark will gather datasets such as `random' and `full' training qualities datasets to bolster training robust CL agents.

\begin{figure*}[h]
% \hspace{-1em}
%\vspace{-0.5em}
 \begin{center}
 \includegraphics[angle=0,width=0.95\textwidth]{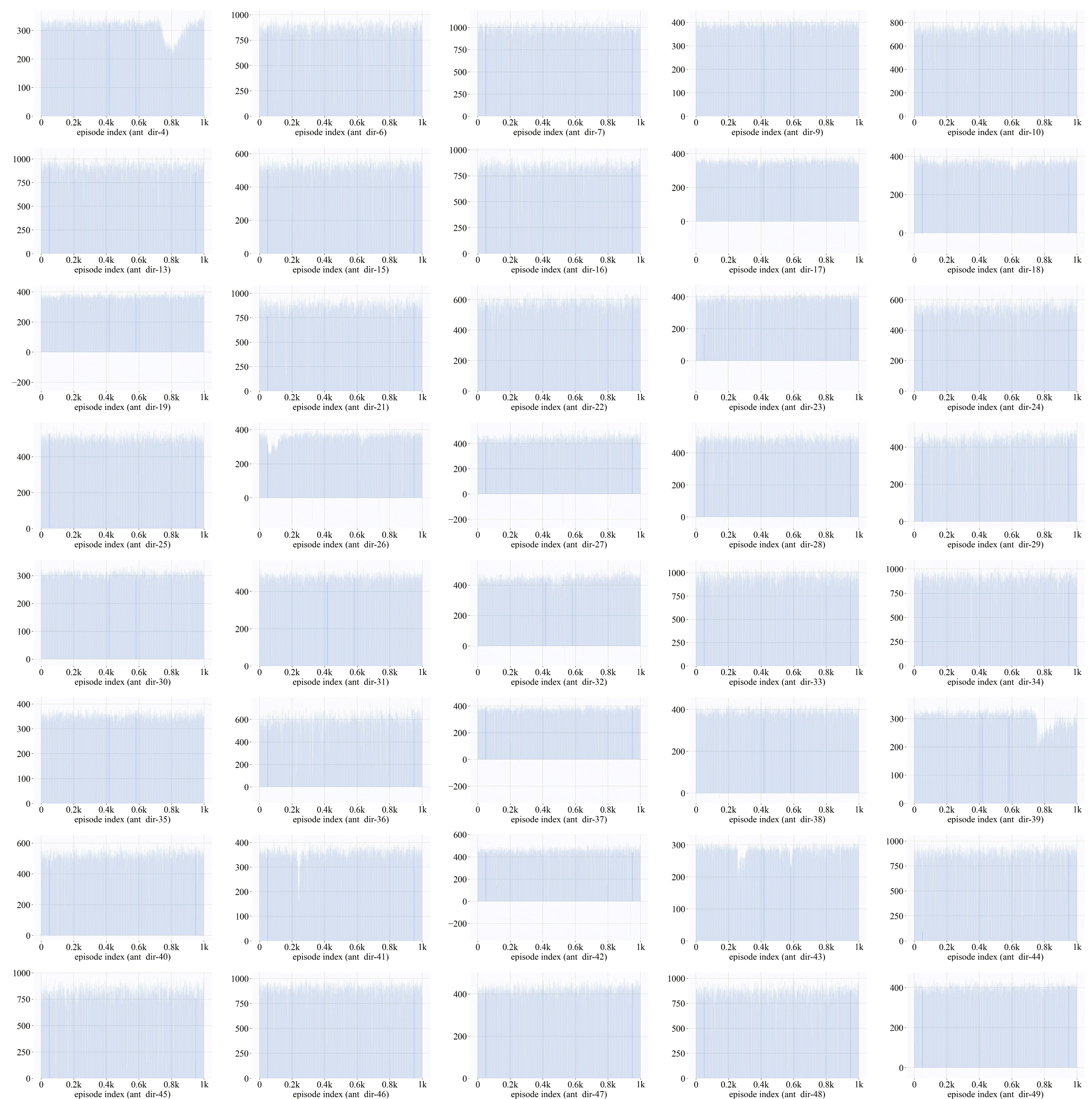}
 \caption{The return statistics of the Ant-dir. We calculate the episode return of the Ant-dir datasets and report the corresponding histogram.}
 \label{ant-dir offline dataset return analysis}
 \end{center}
 \vspace{-0.3cm}
 % \vspace{-0.5em}
 \end{figure*}

Ant-dir is an 8-joint ant environment. 
The different tasks are defined according to the target direction, where the agent should maximize its return with maximal speed in the pre-defined direction. 
As shown in Table~\ref{The information statistics of offline Gym-MuJoCo datasets}, there are 40 tasks (distinguished with ``task id'') with different uniformly sampled goal directions in Ant-dir. 
For each task, the dataset contains approximately 200k transitions, where the observation and action dimensions are 27 and 8, respectively.
We found that the Ant-dir datasets have been used by many researchers~\cite{xu2022prompting, li2020multi, rakelly2019efficient}, so we incorporate them into our benchmark.
Moreover, we report the mean return information of each sub-task in Table~\ref{The information statistics of offline Gym-MuJoCo datasets} and Figure~\ref{ant-dir offline dataset return analysis}.
As for Cheetah-dir, it only contains two tasks that represent forward and backward goal directions.
Compared with Ant-dir, Cheetah-dir possesses lower observation and action space.

\begin{table*}[h]
\centering
% \vspace{-1em}
\small
\caption{The information statistics of offline Continual World-v1 datasets.}
\label{The information statistics of offline continual world datasets}
\resizebox{\textwidth}{!}{
\begin{tabular}{l | r | r | r | r | r | r}
\toprule
\specialrule{0em}{1.5pt}{1.5pt}
\toprule
Continual World dataset & quality & episode length & episode number & mean success & \makecell[c]{observation\\dimension} & \makecell[c]{action\\dimension}\\
\midrule[1pt]
        assembly-v1& expert  & 200                    & 5000           &1.0              & 13                    & 4\\
        basketball-v1& expert  & 200                  & 5000           &1.0              & 13                    & 4\\
        button-press-topdown-v1& expert  & 200        & 5000           &0.99              & 13                    & 4\\
        button-press-topdown-wall-v1& expert  & 200   & 5000           &0.9864              & 13                    & 4\\
        button-press-v1& expert  & 200                & 5000           &0.99              & 13                    & 4\\
        button-press-wall-v1& expert  & 200           & 5000           &1.0              & 13                    & 4\\
        coffee-button-v1& expert  & 200               & 5000           &0.9916              & 13                    & 4\\
        coffee-pull-v1& expert  & 200                 & 5000           &0.99              & 13                    & 4\\
        coffee-push-v1& expert  & 200                 & 5000           &0.99              & 13                    & 4\\
        dial-turn-v1& expert  & 200                   & 5000           &0.9902              & 13                    & 4\\
        disassemble-v1& expert  & 200                 & 5000           &0.0              & 13                    & 4\\
        door-close-v1& expert  & 200                  & 5000           &0.9902              & 13                    & 4\\
        door-open-v1& expert  & 200                   & 5000           &0.9898              & 13                    & 4\\
        drawer-close-v1& expert  & 200                & 5000           &0.9894              & 13                    & 4\\
        drawer-open-v1& expert  & 200                 & 5000           &0.99              & 13                    & 4\\
        faucet-close-v1& expert  & 200                & 5000           &0.9896              & 13                    & 4\\
        faucet-open-v1& expert  & 200                 & 5000           &0.9154              & 13                    & 4\\
        hammer-v1& expert  & 200                      & 5000           &0.99              & 13                    & 4\\
        handle-press-side-v1& expert  & 200           & 5000           &0.9878              & 13                    & 4\\
        handle-press-v1& expert  & 200                & 5000           &0.99              & 13                    & 4\\
        handle-pull-side-v1& expert  & 200            & 5000           &0.9888              & 13                    & 4\\
        handle-pull-v1& expert  & 200                 & 5000           &0.99              & 13                    & 4\\
        lever-pull-v1& expert  & 200                  & 5000           &0.0              & 13                    & 4\\
        peg-insert-side-v1& expert  & 200             & 5000           &0.9604              & 13                    & 4\\
        peg-unplug-side-v1& expert  & 200             & 5000           &0.99              & 13                    & 4\\
        pick-out-of-hole-v1& expert  & 200            & 5000           &0.0              & 13                    & 4\\
        pick-place-v1& expert  & 200                  & 5000           &1.0              & 13                    & 4\\
        pick-place-wall-v1& expert  & 200             & 5000           &0.8196              & 13                    & 4\\
        plate-slide-back-side-v1& expert  & 200       & 5000           &0.99              & 13                    & 4\\
        plate-slide-back-v1& expert  & 200            & 5000           &0.9886              & 13                    & 4\\
        plate-slide-side-v1& expert  & 200            & 5000           &0.7992              & 13                    & 4\\
        plate-slide-v1& expert  & 200                 & 5000           &0.5694              & 13                    & 4\\
        push-back-v1& expert  & 200                   & 5000           &0.9922              & 13                    & 4\\
        push-v1& expert  & 200                        & 5000           &0.9844              & 13                    & 4\\
        push-wall-v1& expert  & 200                   & 5000           &1.00              & 13                    & 4\\
        reach-wall-v1& expert  & 200                  & 5000           &0.99              & 13                    & 4\\
        shelf-place-v1& expert  & 200                 & 5000           &1.00              & 13                    & 4\\
        soccer-v1& expert  & 200                      & 5000           &0.0066              & 13                    & 4\\
        stick-pull-v1& expert  & 200                  & 5000           &0.93              & 13                    & 4\\
        stick-push-v1& expert  & 200                  & 5000           &0.4486              & 13                    & 4\\
        sweep-into-v1& expert  & 200                  & 5000           &0.9662              & 13                    & 4\\
        sweep-v1& expert  & 200                       & 5000           &0.0834              & 13                    & 4\\
        window-close-v1& expert  & 200                & 5000           &0.99              & 13                    & 4\\
        window-open-v1& expert  & 200                 & 5000           &0.99              & 13                    & 4\\
        hammer-v1 & medium  & 200                 & 2000           &0.7689              & 13                    & 4\\
        push-wall-v1 & medium  & 200                 & 2000           &0.7465              & 13                    & 4\\
        faucet-close-v1 & medium  & 200                 & 2000           &0.9364              & 13                    & 4\\
        push-back-v1 & medium  & 200                 & 2000           &0.3168              & 13                    & 4\\
\bottomrule
\specialrule{0em}{1.5pt}{1.5pt}
\bottomrule
\end{tabular}}
\vspace{-0.3cm}
\end{table*}

\begin{table*}[h]
\centering
% \vspace{-1em}
\small
\caption{The information statistics of offline Continual World-v2 datasets.}
\label{The information statistics of offline continual world v2 datasets}
\resizebox{\textwidth}{!}{
\begin{tabular}{l | r | r | r | r | r | r}
\toprule
\specialrule{0em}{1.5pt}{1.5pt}
\toprule
Continual World dataset & quality & episode length & episode number & mean success & \makecell[c]{observation\\dimension} & \makecell[c]{action\\dimension}\\
\midrule[1pt]
        basketball-v2& expert  & 200                  & 5000           &1.0              & 39                    & 4\\
        box-close-v2& expert  & 200   & 5000           &1.0              & 39                    & 4\\
        button-press-topdown-v2& expert  & 200                & 5000           &1.0              & 39                    & 4\\
        button-press-topdown-wall-v2& expert  & 200           & 5000           &1.0              & 39                    & 4\\
        button-press-v2& expert  & 200           & 5000           &1.0              & 39                    & 4\\
        button-press-wall-v2& expert  & 200           & 5000           &1.0              & 39                    & 4\\
        coffee-button-v2& expert  & 200               & 5000           &1.0              & 39                    & 4\\
        dial-turn-v2& expert  & 200               & 5000           &1.0              & 39                    & 4\\
        door-close-v2& expert  & 200                  & 5000           &1.0              & 39                    & 4\\
        door-lock-v2& expert  & 200                   & 5000           &1.0              & 39                    & 4\\
        door-open-v2& expert  & 200                & 5000           &1.0              & 39                    & 4\\
        door-unlock-v2& expert  & 200                 & 5000           &1.0              & 39                    & 4\\
        drawer-close-v2& expert  & 200                 & 5000           &1.0              & 39                    & 4\\
        drawer-open-v2& expert  & 200                 & 5000           &1.0              & 39                    & 4\\
        faucet-close-v2& expert  & 200                & 5000           &1.0              & 39                    & 4\\
        faucet-open-v2& expert  & 200                 & 5000           &1.0              & 39                    & 4\\
        hammer-v2& expert  & 200                      & 5000           &1.0              & 39                    & 4\\
        hand-insert-v2& expert  & 200                      & 5000           &1.0              & 39                    & 4\\
        handle-press-side-v2& expert  & 200           & 5000           &1.0              & 39                    & 4\\
        handle-press-v2& expert  & 200                & 5000           &1.0              & 39                    & 4\\
        handle-pull-side-v2& expert  & 200            & 5000           &1.0              & 39                    & 4\\
        handle-pull-v2& expert  & 200                 & 5000           &1.0              & 39                    & 4\\
        lever-pull-v2& expert  & 200                  & 5000           &1.0              & 39                    & 4\\
        peg-insert-side-v2& expert  & 200             & 5000           &1.0              & 39                    & 4\\
        peg-unplug-side-v2& expert  & 200             & 5000           &1.0              & 39                    & 4\\
        pick-out-of-hole-v2& expert  & 200            & 5000           &1.0              & 39                    & 4\\
        pick-place-v2& expert  & 200                  & 5000           &1.0              & 39                    & 4\\
        plate-slide-back-side-v2& expert  & 200       & 5000           &1.0              & 39                    & 4\\
        plate-slide-back-v2& expert  & 200            & 5000           &1.0              & 39                    & 4\\
        plate-slide-side-v2& expert  & 200            & 5000           &1.0              & 39                    & 4\\
        plate-slide-v2& expert  & 200                 & 5000           &1.0              & 39                    & 4\\
        push-back-v2& expert  & 200                   & 2668           &1.0              & 39                    & 4\\
        push-v2& expert  & 200                        & 5000           &1.0              & 39                    & 4\\
        push-wall-v2& expert  & 200                   & 5000           &1.0              & 39                    & 4\\
        reach-v2& expert  & 200                  & 5000           &1.0              & 39                    & 4\\
        reach-wall-v2& expert  & 200                  & 5000           &1.0              & 39                    & 4\\
        shelf-place-v2& expert  & 200                 & 5000           &1.0              & 39                    & 4\\
        stick-pull-v2& expert  & 200                  & 5000           &1.0              & 39                    & 4\\
        stick-push-v2& expert  & 200                  & 5000           &1.0              & 39                    & 4\\
        sweep-into-v2& expert  & 200                  & 5000           &1.0              & 39                    & 4\\
        sweep-v2& expert  & 200                       & 5000           &1.0              & 39                    & 4\\
        window-close-v2& expert  & 200                & 5000           &1.0              & 39                    & 4\\
        window-open-v2& expert  & 200                 & 5000           &1.0              & 39                    & 4\\
\bottomrule
\specialrule{0em}{1.5pt}{1.5pt}
\bottomrule
\end{tabular}}
\vspace{-0.3cm}
\end{table*}

\begin{table*}[h]
\centering
% \vspace{-1em}
\small
\caption{The information statistics of offline Gym-MuJoCo datasets.}
\label{The information statistics of offline Gym-MuJoCo datasets}
\resizebox{\textwidth}{!}{
\begin{tabular}{l | r | r | r | r | r | r | r}
\toprule
\specialrule{0em}{1.5pt}{1.5pt}
\toprule
MuJoCo dataset & task id & quality & episode length & episode number & mean return & \makecell[c]{observation\\dimension} & \makecell[c]{action\\dimension}\\
\midrule[1pt]
        Ant-dir& 4 & expert  & 200            & 999           &315.7402              & 27                    & 8\\
        Ant-dir& 6 & expert  & 200            & 1000           &865.8379              & 27                    & 8\\
        Ant-dir& 7 & expert  & 200            & 1000           &993.9981              & 27                    & 8\\
        Ant-dir& 9 & expert  & 200            & 999           &390.8016              & 27                    & 8\\
        Ant-dir& 10 & expert  & 200            & 1000           &744.8206              & 27                    & 8\\
        Ant-dir& 13 & expert  & 200            & 1000           &922.9069              & 27                    & 8\\
        Ant-dir& 15 & expert  & 200            & 1000           &522.9190              & 27                    & 8\\
        Ant-dir& 16 & expert  & 200            & 1000           &835.9635              & 27                    & 8\\
        Ant-dir& 17 & expert  & 200            & 999           &352.7341              & 27                    & 8\\
        Ant-dir& 18 & expert  & 200            & 1000           &367.9050              & 27                    & 8\\
        Ant-dir& 19 & expert  & 200            & 999           &369.9799              & 27                    & 8\\
        Ant-dir& 21 & expert  & 200            & 1000           &868.7162              & 27                    & 8\\
        Ant-dir& 22 & expert  & 200            & 1000           &577.2005              & 27                    & 8\\
        Ant-dir& 23 & expert  & 200            & 1000           &386.7926              & 27                    & 8\\
        Ant-dir& 24 & expert  & 200            & 1000           &547.0642              & 27                    & 8\\
        Ant-dir& 25 & expert  & 200            & 1000           &501.6898              & 27                    & 8\\
        Ant-dir& 26 & expert  & 200            & 1000           &357.3981              & 27                    & 8\\
        Ant-dir& 27 & expert  & 200            & 1000           &439.8590              & 27                    & 8\\
        Ant-dir& 28 & expert  & 200            & 1000           &484.8640              & 27                    & 8\\
        Ant-dir& 29 & expert  & 200            & 1000           &439.0989              & 27                    & 8\\
        Ant-dir& 30 & expert  & 200            & 999           &305.6620              & 27                    & 8\\
        Ant-dir& 31 & expert  & 200            & 999           &478.8927              & 27                    & 8\\
        Ant-dir& 32 & expert  & 200            & 999           &442.5488              & 27                    & 8\\
        Ant-dir& 33 & expert  & 200            & 1000           &952.0699              & 27                    & 8\\
        Ant-dir& 34 & expert  & 200            & 1000           &909.5234              & 27                    & 8\\
        Ant-dir& 35 & expert  & 200            & 999           &352.6703              & 27                    & 8\\
        Ant-dir& 36 & expert  & 200            & 1000           &593.1572              & 27                    & 8\\
        Ant-dir& 37 & expert  & 200            & 1000           &374.4446              & 27                    & 8\\
        Ant-dir& 38 & expert  & 200            & 999           &390.5748              & 27                    & 8\\
        Ant-dir& 39 & expert  & 200            & 999           &307.2525              & 27                    & 8\\
        Ant-dir& 40 & expert  & 200            & 1000           &524.2991              & 27                    & 8\\
        Ant-dir& 41 & expert  & 200            & 1000           &360.6967              & 27                    & 8\\
        Ant-dir& 42 & expert  & 200            & 1000           &454.5446              & 27                    & 8\\
        Ant-dir& 43 & expert  & 200            & 999           &285.9895              & 27                    & 8\\
        Ant-dir& 44 & expert  & 200            & 1000           &878.4141              & 27                    & 8\\
        Ant-dir& 45 & expert  & 200            & 1000           &813.5594              & 27                    & 8\\
        Ant-dir& 46 & expert  & 200            & 1000           &900.4641              & 27                    & 8\\
        Ant-dir& 47 & expert  & 200            & 1000           &422.5884              & 27                    & 8\\
        Ant-dir& 48 & expert  & 200            & 1000           &865.0776              & 27                    & 8\\
        Ant-dir& 49 & expert  & 200            & 1000           &398.1321              & 27                    & 8\\
        Cheetah-dir& 0 & expert  & 200            & 999           &666.5849              & 20                    & 6\\
        Cheetah-dir& 1 & expert  & 200            & 999           &1134.3012              & 20                    & 6\\
\bottomrule
\specialrule{0em}{1.5pt}{1.5pt}
\bottomrule
\end{tabular}}
\vspace{-0.3cm}
\end{table*}

\end{document}